\documentclass[10pt,twocolumn,letterpaper]{article}
\usepackage{cvpr}
\usepackage[utf8]{inputenc}
\usepackage{graphicx}
\usepackage{color}
\usepackage{times}
\usepackage{amsmath,mathtools}
\usepackage{amsfonts}
\usepackage{amsthm}	
\usepackage{amssymb}
\usepackage{enumerate}
\usepackage{algorithm} 
\usepackage{algpseudocode}
\usepackage{booktabs}
\usepackage{tabularx}
\usepackage{array}
\usepackage{bm}
\usepackage{url}
\usepackage{bbm}
\usepackage{enumitem}
\usepackage{stackengine}
\usepackage[algo2e,inoutnumbered,linesnumbered,algoruled,vlined]{algorithm2e}
\usepackage{xcolor}
\usepackage{multirow}
\usepackage{authblk}
\usepackage{units}
\usepackage{algorithm}
\usepackage{algpseudocode}
\usepackage{gensymb}
\usepackage[nodisplayskipstretch]{setspace}
\usepackage{mathtools, nccmath}
\usepackage[caption=false]{subfig}

\definecolor{bostonuniversityred}{rgb}{0.8, 0.0, 0.0}
\newcommand{\rev}[1]{\textcolor{black}{#1}}

\usepackage[pagebackref=true,breaklinks=true,letterpaper=true,colorlinks,bookmarks=false]{hyperref}

\cvprfinalcopy %
\usepackage{cleveref}

\def\cvprPaperID{1558} %

\ifcvprfinal\fi

\newcommand{\Rot}{\mathbf{R}}

\newcommand{\M}{\mathbf{M}}

\newcommand{\sing}{\mathbf{S}}

\newcommand{\R}{\mathbb{R}}

\DeclareMathOperator*{\argmin}{arg\min}

\crefname{section}{§}{§§}
\Crefname{section}{§}{§§}

\crefname{section}{§}{§§}
\Crefname{section}{§}{§§}
\crefname{thm}{Thm.}{Thm.}
\crefname{eq}{Eq.}{Eq.}
\crefname{figure}{Fig.}{Fig.}
\crefname{dfn}{Dfn.}{Dfn.}
\crefname{table}{Tab.}{Tab.}
\begin{document}

\title{\vspace{-0.45cm}Learning multiview 3D point cloud registration}
\author{\vspace{-25pt}Zan Gojcic$^{*\mathsection}$ \qquad Caifa Zhou$^{*\mathsection}$ \qquad Jan D. Wegner$^{\mathsection}$ \qquad Leonidas J. Guibas$^\dag$ \qquad Tolga Birdal$^\dag$ \\
$^{\mathsection}$ETH Zurich \qquad $^\dag$Stanford University}
\maketitle
\begin{abstract}
We present a novel, end-to-end learnable, multiview 3D point cloud registration algorithm. Registration of multiple scans typically follows a two-stage pipeline: the initial pairwise alignment and the globally consistent refinement. The former is often ambiguous due to the low overlap of neighboring point clouds, symmetries and repetitive scene parts. Therefore, the latter global refinement aims at establishing the cyclic consistency across multiple scans and helps in resolving the ambiguous cases. In this paper we propose, to the best of our knowledge, the first end-to-end algorithm for joint learning of both parts of this two-stage problem. Experimental evaluation on well accepted benchmark datasets shows that our approach outperforms the state-of-the-art by a significant margin, while being end-to-end trainable and computationally less costly. Moreover, we present detailed analysis and an ablation study that validate the novel components of our approach. The source code and pretrained models are publicly available under \url{https://github.com/zgojcic/3D_multiview_reg}.
\end{abstract}

\section{Introduction}
{\let\thefootnote\relax\footnote{{$^*$First two authors contributed equally to this work.}}}
Downstream tasks in 3D computer vision, such as semantic segmentation and object detection typically require a holistic representation of the scene. The capability of aligning and fusing individual point cloud fragments, which cover only small parts of the environment, into a globally consistent holistic representation is therefore essential and has several use cases in augmented reality and robotics. Pairwise registration of adjacent fragments is a well studied problem and traditional approaches based on geometric constraints~\cite{rabbani2007,zeisl2013,theiler2015} and hand-engineered feature descriptors~\cite{johnson1999,flint2007,rusu2009FPFH,tombari2010SHOT} have shown successful results to some extent. Nevertheless, in the recent years, research on local descriptors for pairwise registration of 3D point clouds is centered on deep learning approaches~\cite{zeng20163dmatch,khoury2017CGF,deng2018ppfnet,yew20183dfeatnet,Deng2018PPFFoldNetUL,gojcic20193DSmoothNet} that succeed in capturing and encoding evidence hidden to hand-engineered descriptors. Furthermore, novel end-to-end methods for pairwise point cloud registration were recently proposed~\cite{Wang_2019_DCP,Lu_2019_DeepVCP}. While demonstrating good performance for many tasks, pairwise registration of individual views of a scene has some conceptual drawbacks: (i) low overlap of adjacent point clouds can lead to inaccurate or wrong matches, (ii) point cloud registration has to rely on very local evidence, which can be harmful if 3D scene structure is scarce or repetitive, (iii) separate post-processing is required to combine all pair-wise matches into a global representation.
Compared to the pairwise methods, globally consistent multiview alignment of unorganized point cloud fragments is yet to fully benefit from the recent advances achieved by the deep learning methods. State-of-the art methods typically still rely on a good initialization of the pairwise maps, which they try to refine globally in a subsequent decoupled step~\cite{ govindu2004liesync,torsello2011multiview,arie2012global,arrigoni2014robust,bernard2015MAtranssync,arrigoni2016se3sync,maset2017practical, birdal2018bayesian}. A general drawback of this hierarchical procedure is that global noise distribution over all nodes of the pose graph ends up being far from random, i.e. significant biases persist due to the highly correlated initial pairwise maps.

\begin{figure}
    \centering
    \includegraphics[width=\linewidth]{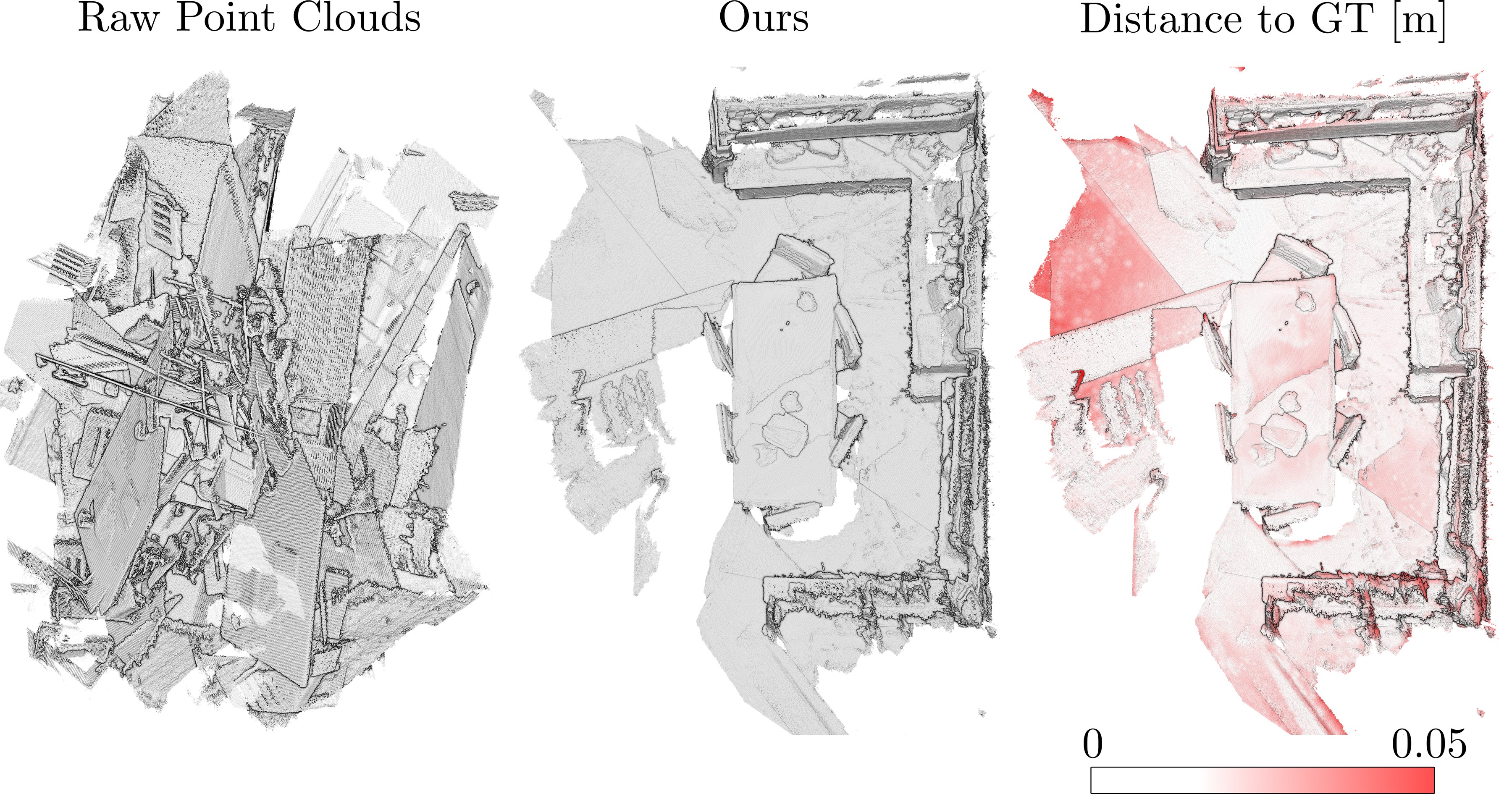}
    \caption{Result of our end-to-end reconstruction on the 60 scans of \textit{Kitchen} scene from \textit{3DMatch benchmark}~\cite{zeng20163dmatch}.}
    \label{fig:teaser}\vspace{-5mm}
\end{figure}
In this paper, we present, to the best of our knowledge, the first \emph{end-to-end data driven multiview point cloud registration algorithm}. Our method takes a set of potentially overlapping point clouds as input and outputs a global/absolute transformation matrix per each of the input scans (\cf Fig.~\ref{fig:teaser}). We depart from a traditional two-stage approach where the individual stages are detached from each other and directly learn to register all views of a scene in a globally consistent manner.

The main contributions of our work are:
\begin{itemize}
    %\item Building on \cite{Choy2019FCGF} and \cite{zhang2019oanet} we propose a novel end-to-end pairwise 3D point cloud registration algorithm.
    \item We formulate the traditional two-stage approach in an end-to-end neural network, which in the forward pass solves two differentiable optimization problems: (i) the Procrustes problem for the estimation of the pairwise transformation parameters and (ii) the spectral relaxation of the transformation synchronization.
    
    \item We propose a confidence estimation block that uses a novel \textit{overlap pooling} layer to predict the confidence in the estimated pairwise transformation parameters.
    
    \item We cast the mutliview 3D point cloud registration problem as an iterative reweighted least squares (IRLS) problem and iteratively refine both the pairwise and absolute transformation estimates.
\end{itemize}
Resulting from the aforementioned contributions, the proposed multiview registration algorithm (i) is very efficient to compute, (ii) achieves more accurate scan alignments because the residuals are being fed back to the pairwise network in an iterative manner, (iii) outperforms current state-of-the-art on pairwise as well as multiview point cloud registration.

\section{Related Work}
\paragraph{Pairwise registration}
The traditional pairwise registration pipeline consists of two stages: the coarse alignment stage, which provides the initial estimate of the relative transformation parameters and the refinement stage that iteratively refines the transformation parameters by minimizing the 3D registration error under the assumption of rigid transformation. 

The former is traditionally performed by using either handcrafted~\cite{rusu2009FPFH, tombari2010SHOT, tombari2010USC} or learned~\cite{zeng20163dmatch,khoury2017CGF,deng2018ppfnet,deng2018ppffold,yew20183dfeatnet, gojcic20193DSmoothNet,Choy2019FCGF} 3D local features descriptors to establish the pointwise candidate correspondences in combination with a RANSAC-like robust estimator~\cite{fischer1981RANSAC,raguram2012usac, korman2018latent} or geometric hashing~\cite{drost2010model,birdal2015point, hinterstoisser2016going}. A parallel stream of works~\cite{aiger20084PCS, theiler2014keypoint,mellado2014super} relies on establishing correspondences using the 4-point congruent sets. 
In the refinement stage, the coarse transformation parameters are often fine-tuned with a variant of the iterative closest point (ICP) algorithm~\cite{besl1992ICP}. ICP-like algorithms~\cite{li20073d, yang2015go} perform optimization by alternatively hypothesizing the correspondence set and estimating the new set of transformation parameters. They are known to not be robust against outliers and to converge to a global optimum only when starting with a good prealingment~\cite{birdal2017cad}. ICP algorithms are often extended to use additional radiometric, temporal or odometry constraints~\cite{zhou2016FGR}. Contemporary to our work,~\cite{Wang_2019_DCP,Lu_2019_DeepVCP} propose to integrate coarse and fine pairwise registration stages into an end-to-end learnable algorithm. \rev{Using a deep network,~\cite{gross2019alignnet} formulates the object tracking as a relative motion estimation of two point sets.}

\vspace{2mm}\noindent\textbf{Multiview registration\,}
Multiview, global point cloud registration methods aim at resolving hard or ambiguous cases that arise in pairwise methods by incorporating cues from multiple views. The first family of methods employ a multiview ICP-like scheme to optimize for camera poses as well as 3D point correspondences~\cite{huber2003fully,fantoni2012accurate,mian2006three,birdal2017cad}. A majority of these suffer from increased complexity of correspondence estimation. To alleviate this, some approaches only optimize for motion and use the scans to evaluate the registration error~\cite{zhou2016FGR,theiler2015,Bhattacharya2019robustReg}. Taking a step further, other modern methods make use of the global cycle-consistency and optimize only over the poses starting from an initial set of pairwise maps. This efficient approach is known as \emph{synchronization}~\cite{birdal2019probabilistic,torsello2011multiview,arie2012global,theiler2015,arrigoni2014robust,bernard2015MAtranssync,maset2017practical,zhou2016FGR,Bhattacharya2019robustReg,huang2019learningTransSync}.
Global structure-from-motion~\cite{cui2015globalsfm,zhu2018gsfm} aims to synchronize the observed relative motions by decomposing rotation, translation and scale components.~\cite{ding2019} proposes a global point cloud registration approach using two networks, one for pose estimation and another modelling the scene structure by estimating the occupancy status of global coordinates.

Probably the most similar work to ours is~\cite{huang2019learningTransSync}, where the authors aim to adapt the edge weights for the transformation synchronization layer by learning a data driven weighting function. %
A major conceptual difference to our approach is that relative transformation parameters are estimated using FPFH~\cite{rusu2009FPFH} in combination with FGR~\cite{zhou2016FGR} and thus, unlike ours, are not learned. Furthermore, in each iteration~\cite{huang2019learningTransSync} has to convert the point clouds to depth images as the weighting function is approximated by a 2D CNN. On the other hand our whole approach operates directly on point clouds, is fully differentiable and therefore facilitates learning a global, multiview point cloud registration in an end-to-end manner.

\section{End-to-End Multiview 3D Registration}
\label{sec:methodology}
%In this section we describe all components of our multiview 3D registration algorithm. 
In this section we derive the proposed multiview 3D registration algorithm as a composition of functions depending upon the data. The network architectures used to approximate these functions are then explained in detail in Sec~\ref{sec:network_architecture}. We begin with a new algorithm for learned pairwise point cloud registration, which uses two point clouds as input and outputs estimated transformation parameters (Sec.~\ref{sec:pairwise_registration_theory}). This method is extended to multiple point clouds by using a transformation synchronization layer amenable to backpropagation (Sec.~\ref{sec:transform_sync_theory}). %The input graph to this synchronization layer is learned by a neural network, which encodes, along with the relative transformation parameters, the confidence of pairwise point cloud matches as edge information in the global scene graph.
\rev{The input graph to this synchronization layer encodes, along with the relative transformation parameters, the confidence in these pairwise maps, which is also estimated using a novel neural network, as edge information}. Finally, we \rev{propose} an \rev{IRLS} scheme (Sec.~\ref{sec:iterative_refinment}) to refine the global registration of all point clouds by updating the edge weights as well as the pairwise poses.

Consider a set of potentially overlapping point clouds $\mathcal{S} = \{\sing_i \in \R^{N\times 3},1\leq i \leq N_\mathcal{S}\}$ capturing a 3D scene from different viewpoints (i.e. poses). The task of \textit{multiview registration} is to recover the rigid, absolute poses $\{\mathbf{M}^*_i\in SE(3)\}_i$ given the scan collection, where
\begin{equation}
\label{eq:se3}
SE(3) = \left \{
\mathbf{M} \in \R^{4\times 4} \colon \mathbf{M}=\begin{bmatrix} 
\Rot & \mathbf{t} \\ 
\mathbf{0}^\top & 1 
\end{bmatrix}
\right\},
\end{equation}
$\mathbf{R}_i \in SO(3)$ and $\mathbf{t_i} \in \mathbb{R}^3$. $\mathcal{S}$ can be augmented by connectivity information resulting in a finite graph $\mathcal{G} = (\mathcal{S},\mathcal{E})$, where each vertex represents a single point set and the edges $(i,j) \in \mathcal{E}$ encode the information about the relative rotation $\mathbf{R}_{ij}$ and translation $\mathbf{t}_{ij}$ between the vertices. These relative transformation parameters satisfy $\mathbf{R}_{ij} = \mathbf{R}^T_{ji}$ and $\mathbf{t}_{ij} = -\mathbf{R}^T_{ij}\mathbf{t}_{ji}$ as well as the \textit{compatibility constraint}~\cite{arrigoni2016se3sync}
\begin{equation}
    \mathbf{R}_{ij} \approx \mathbf{R}_i\mathbf{R}_j^T \quad\quad\quad \mathbf{t}_{ij} \approx -\mathbf{R}_i\mathbf{R_j}^T\mathbf{t}_j + \mathbf{t}_i
\end{equation}
In current state-of-the-art~\cite{zhou2016FGR,huang2019learningTransSync,Bhattacharya2019robustReg} edges $\mathcal{E}$ of $\mathcal{G}$ are initialized with (noisy) relative transformation parameters $\{\M_{ij}\}$, obtained by an independent, auxiliary pairwise registration algorithm. Global scene consistency is enforced via a subsequent synchronization algorithm. In contrast, we propose a joint approach where pairwise registration and transformation synchronization are tightly coupled as one fully differentiable component, which leads to an end-to-end learnable, global registration pipeline. 
%\vspace{-5mm}
\subsection{Pairwise registration of point clouds}
\label{sec:pairwise_registration_theory}
\begin{figure*}
    \centering
    \includegraphics[width=\linewidth]{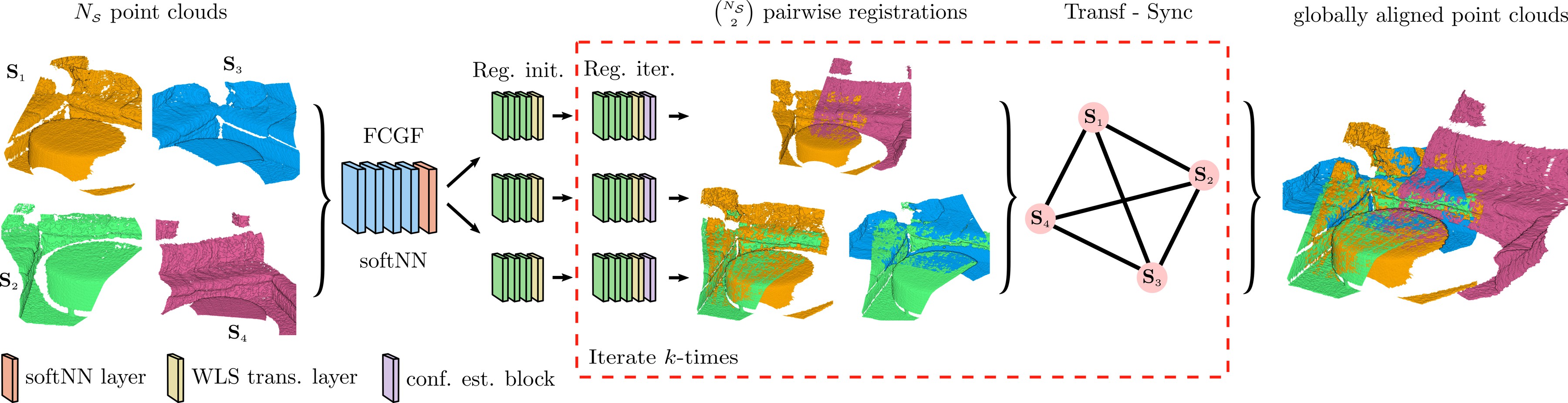}
    \caption{Proposed pipeline for end-to-end multiview 3D point cloud registration. For each of the input point clouds $\mathbf{S}_i$ we extract FCGF~\cite{Choy2019FCGF} features that are fed to the softNN layer to compute the stochastic correspondences for $N_{\mathcal{S}} \choose 2$ pairs. These correspondences are used as input to the initial registration block (i.e. \textit{Reg. init.}) that outputs the per-correspondence weights, initial transformation parameters, and per-point residuals. Along with the correspondences, the initial weights and residuals are then input to the registration refinement block (i.e. \textit{Reg. iter.}), whose outputs are used to build the graph. After each iteration of \rev{the} \textit{Transf-Sync} layer the estimated transformation parameters are used to pre-align the correspondences that are concatenated with the weights from the previous iteration and the residuals and feed anew to \textit{Reg. iter.} block. We iterate over the \textit{Reg. iter.} and \textit{Transf-Sync} layer for four times.}
    \label{fig:network_architecture}\vspace{-3mm}
\end{figure*}
In the following, we introduce a differentiable, pairwise registration algorithm that can easily be incorporated into an end-to-end multiview 3D registration algorithm. %
Let $\{\mathbf{P},\mathbf{Q}\}:=\{\mathbf{S}_i,\mathbf{S}_j | i\neq j \} \subset \mathcal{S}$ denote a pair of point clouds where $(\mathbf{P})_l =: \mathbf{p}_l\in \mathbb{R}^3$ and $(\mathbf{Q})_l =: \mathbf{q}_l\in \mathbb{R}^3$ represent the coordinate vectors of individual points in point clouds $\mathbf{P} \in \mathbb{R}^{N_\mathbf{P} \times 3}$ and $\mathbf{Q}  \in \mathbb{R}^{N_\mathbf{Q} \times 3}$, respectively. The goal of pairwise registration is to retrieve optimal $\hat{\mathbf{R}}_{ij}$ and $\hat{\mathbf{t}}_{ij}$.%
\begin{equation}
    \hat{\mathbf{R}}_{ij},\hat{\mathbf{t}}_{ij} = \argmin_{\mathbf{R}_{ij},\mathbf{t}_{ij}} \sum_{l=1}^{N_{\mathbf{P}}}|| \mathbf{R}_{ij}\mathbf{p}_l + \mathbf{t}_{ij} - \phi({\mathbf{p}_l,\mathbf{Q}}) ||^2
    \label{eq:pairwise}
\end{equation}
where $\phi({\mathbf{p},\mathbf{Q}})$ is a \textit{correspondence function} that maps the points $\mathbf{p}$ to their corresponding points in point cloud $\mathbf{Q}$. The formulation of Eq.~\ref{eq:pairwise} facilitates a differentiable closed-form solution, which is---subject to the noise distribution---close to the ground truth solution \cite{sorkine2017least}. However, least square solutions are not robust and thus Eq.~\ref{eq:pairwise} will yield wrong transformation parameters in case of high outlier ratio. In practice, the mapping $\phi({\mathbf{p},\mathbf{Q}})$ is far from ideal and erroneous correspondences typically dominate. To circumvent that, Eq.~\ref{eq:pairwise} can be robustified against outliers by introducing a heteroscedastic weighting matrix~\cite{torr1997development,sorkine2017least}:
\begin{equation}
    \hat{\mathbf{R}}_{ij}, \hat{\mathbf{t}}_{ij} = \argmin_{\mathbf{R}_{ij},\mathbf{t}_{ij}} \sum_{l=1}^{N_{\mathbf{P}}} w_l|| \mathbf{R}_{ij}\mathbf{p}_l + \mathbf{t}_{ij} - \phi({\mathbf{p}_l,\mathbf{Q}}) ||^2
    \label{eq:weighted_pairwise}
\end{equation}
where $w_l := (\mathbf{w})_l$ is the weight of the putative correspondence $\gamma_l \in \mathbb{R}^6 = \{\mathbf{p}_l,\phi({\mathbf{p}_l,\mathbf{Q}})\}$ computed by some weighting function $\mathbf{w} = \psi_{\text{init}}(\bm{\Gamma})$, where $\bm{\Gamma} := \{\gamma_l\} := \{\mathbf{P},\{\phi({\mathbf{p}_l,\mathbf{Q}})\}_l\}$ and $\psi_{\text{init}}: \mathbb{R}^{N_\mathbf{P} \times 6} \mapsto \ \mathbb{R}^{N_\mathbf{P}} $. Assuming that $w_l$ is close to one when the putative correspondence is an inlier and close to zero otherwise,  Eq.~\ref{eq:weighted_pairwise} will yield the correct transformation parameters while retaining a differentiable closed-form solution~\cite{sorkine2017least}. Hereinafter we denote this closed-form solution as weighted least squares transformation \textit{WLS trans.}~and for the sake of completeness, its derivation is provided in the supp.~material.
%\vspace{-2mm}
\subsection{Differentiable transformation synchronization}
\label{sec:transform_sync_theory}
Returning to the task of multiview registration, we again consider the initial set of point clouds $\mathcal{S}$. If no prior connectivity information is given, graph $\mathcal{G}$ can be initialized by forming ${N_\mathcal{S} \choose 2}$ point cloud pairs and estimating their relative transformation parameters as described in Sec.~\ref{sec:pairwise_registration_theory}. The global transformation parameters can be estimated either jointly (\textit{transformation synchronization})~\cite{govindu2004liesync, bernard2015MAtranssync, arrigoni2016se3sync, birdal2018bayesian} or by dividing the problem into \textit{rotation synchronization}~\cite{arie2012global,arrigoni2014robust} and \textit{translation synchronization}~\cite{huang2017transSync}. \rev{Herein, we opt for the latter approach, which under the spectral relation admits a differentiable closed-form solution~\cite{arie2012global,arrigoni2014robust,huang2017transSync}.}

\vspace{2mm}\noindent\textbf{Rotation synchronization\,}
The goal of rotation synchronization is to retrieve global rotation matrices $\{\mathbf{R}_i^*\}$ by solving the following minimization problem based on their observed ratios $\{\hat{\mathbf{R}}_{ij}\}$ 
\begin{equation}
    \mathbf{R}_i^* = \argmin_{\mathbf{R_i} \in SO(3)} \sum_{(i,j) \in \mathcal{E}} c_{ij} ||\hat{\mathbf{R}}_{ij} - \mathbf{R}_i\mathbf{R}_j^T||^2_F
    \label{eq:rot_sync}
\end{equation}
where the weigths $c_{ij} := \zeta_{\text{init}}(\bm{\Gamma})$ represent the confidence in the relative transformation parameters $\hat{\mathbf{M}}_{ij}$. Under the spectral relaxation Eq.~\ref{eq:rot_sync} admits a closed-form solution, which is provided in the supp.~material~\cite{arie2012global, arrigoni2014robust}.

%Consider a symmetric matrix of relative transformations $\mathbf{L} = \mathbf{D} - \mathbf{A}$~\cite{singer2012vector}, where $ \mathbf{D}  \in  \mathbb{R}^{3N_\mathcal{S}  \times 3N_\mathcal{S}}$ is a diagonal degree matrix and $\mathbf{A} \in  \mathbb{R}^{3N_\mathcal{S}  \times 3N_\mathcal{S}}$ is constructed as
%\begin{equation}
%    \mathbf{A}  = \begin{bmatrix*}[c] 
%    \mathbf{0}_3 & c_{12}\hat{\mathbf{R}}_{12} & \cdots & %c_{1N_\mathcal{S}}\hat{\mathbf{R}}_{1N_\mathcal{S}}\\
%    c_{21}\hat{\mathbf{R}}_{21} & \mathbf{0}_3 &  \cdots & %c_{2N_\mathcal{S}}\hat{\mathbf{R}}_{2N_\mathcal{S}}\\
%    \vdots &  & \ddots & \vdots\\
%    c_{N_\mathcal{S}1}\hat{\mathbf{R}}_{N_\mathcal{S}1} &  %c_{N_\mathcal{S}2}\hat{\mathbf{R}}_{N_\mathcal{S}2} & \cdots & \mathbf{0}_3 \\
%\end{bmatrix*}
%\end{equation}
%where the weigths $c_{ij} := \zeta_{\text{init}}(\bm{\Gamma})$ represent the confidence in the relative transformation parameters $\hat{\mathbf{M}}_{ij}$. The least squares estimates of the global rotation matrices $\mathbf{R}^*_i$ are then given, under relaxed orthonormality and determinant constraint, by the three eigenvectors $\mathbf{v}_i \in \mathbb{R}^{3N_\mathcal{S}}$ corresponding to the smallest eigenvalues of $\mathbf{L}$. Consequently, the nearest rotation matrices under Frobenius norm can be obtained by a projection of the $3\times3$ submatrices of $\mathbf{V}=[\mathbf{v}_1,\mathbf{v}_2,\mathbf{v}_3] \in \mathbb{R}^{3N_\mathcal{S}\times 3}$ onto the orthonormal matrices and enforcing the $\det(\mathbf{R}_i^*)=1$ (to avoid the reflections).
\vspace{-2mm}
\paragraph{Translation synchronization}
Similarly, the goal of translation synchronization is to retrieve global translation vectors $\{\mathbf{t}_i^*\}$ that minimize the following least squares problem
\begin{equation}
    \mathbf{t}_i^* = \argmin_\mathbf{t_i} \sum_{(i,j) \in \mathcal{E}} c_{ij}||\hat{\mathbf{R}}_{ij}\mathbf{t}_{i} + \hat{\mathbf{t}}_{ij} - \mathbf{t}_{j} ||^2
    \label{eq:trans_sync}
\end{equation}
The differentiable closed-form solution to Eq.~\ref{eq:trans_sync} is again provided in the supp.~material.

%Considering the normalized version $\mathbf{L}_{\text{norm.}}$ of the above defined matrix of relative transformations $\mathbf{L}$, the closed-form solution of Eq.~\ref{eq:trans_sync} can be written as~\cite{huang2019learningTransSync}
%\begin{equation}
%    \mathbf{t}^* = \mathbf{L}_{\text{norm.}}^+\mathbf{b}
%\end{equation}
%where $\mathbf{t}^* = [\mathbf{t}_1^{*^{T}},\dots,\mathbf{t}_{N_{\mathcal{S}}}^{*^{T}}]^T \in \mathbb{R}^{3N_\mathcal{S}}$ and $\mathbf{b} = [\mathbf{b}_1^{*^{T}},\dots,\mathbf{b}_{N_{\mathcal{S}}}^{*^{T}}]^T \in \mathbb{R}^{3N_\mathcal{S}}$ with
%\begin{equation}
%    \mathbf{b}_i := -\hspace{-0.1ex}\sum_{j \in \mathcal{N}(i)} %w_{ij}\hat{\mathbf{t}}_{ij}\text{ and }\mathbf{b}_j := %-\hspace{-0.1ex}\sum_{i \in \mathcal{N}(j)} w_{ij} %\hat{\mathbf{R}}^T_{ij}\hat{\mathbf{t}}_{ij}.
%\end{equation}
%where $\mathcal{N}(i)$ ($\mathcal{N}(j)$) denotes all the neighboring %vertices of $\mathbf{S}_i$ ($\mathbf{S}_j$) in graph $\mathcal{G}$.
%\vspace{-3mm}
\subsection{Iterative refinement of the registration}
\label{sec:iterative_refinment}
The above formulation (Sec.~\ref{sec:pairwise_registration_theory} and~\ref{sec:transform_sync_theory}) facilitates an implementation in an iterative scheme, which in turn can be viewed as an IRLS algorithm. We can start each subsequent iteration $(k+1)$ by pre-aligning the point cloud pairs using the synchronized estimate of the relative transformation parameters $\mathbf{M}^{*(k)}_{ij} = \mathbf{M}_{i}^{*(k)}\mathbf{M}_{j}^{*(k)^{-1}}$ from iteration $(k)$ such that $\mathbf{Q}^{(k+1)} := \mathbf{M}^{*(k)}_{ij} \otimes \mathbf{Q}$, where $\otimes$ denotes applying the transformation $\mathbf{M}^{*(k)}_{ij}$ to point cloud $\mathbf{Q}$. Additionally, weights $\mathbf{w}^{(k)}$ and residuals $\mathbf{r}^{(k)}$ of the previous iteration can be used as a side information in the correspondence weighting function. Therefore, $\psi_{\text{init}}(\cdot)$ is extended to
%\begin{equation}
%    \mathbf{w}^{(k+1)}:= \psi_{\text{iter}}(\{\mathbf{P},\phi(\mathbf{P},\mathbf{Q}^{(k+1)})\},\mathbf{w}^{(k)},\mathbf{r}^{(k)}),
%\end{equation}
\begin{equation}
    \mathbf{w}^{(k+1)}:= \psi_{\text{iter}}(\bm{\Gamma}^{(k + 1)},\mathbf{w}^{(k)},\mathbf{r}^{(k)}),
\end{equation}
where $\bm{\Gamma}^{(k+1)} := \{\gamma_l^{(k+1)}\} := \{\mathbf{P},\{\phi({\mathbf{p}_l,\mathbf{Q}^{(k+1)}})\}_l\}$.
Analogously, the difference between the input $\hat{\mathbf{M}}^{(k)}_{ij}$ and the synchronized $\mathbf{M}^{*(k)}_{ij}$ transformation parameters of the $(k)\mathrm{-th}$ iteration can be used as an additional cue for estimating the confidence $c^{(k+1)}_{ij}$. Thus, $\zeta_{\text{init}}(\cdot)$ can be extended to
\begin{equation}
    c^{(k+1)}_{ij} := \zeta_{\text{iter}}(\bm{\Gamma}^{(k+1)},\hat{\mathbf{M}}^{(k)}_{ij},\mathbf{M}^{*(k)}_{ij} ).
\end{equation}

\section{Network Architecture}
\label{sec:network_architecture}
%We implement our proposed \textit{multiview registration} algorithm as a deep neural network (Fig.~\ref{fig:network_architecture}). In this section, we first describe all individual components (each representing one component of the algorithm presented in Sec.~\ref{sec:methodology}) before integrating them into one fully differentiable, end-to-end trainable algorithm. 
We implement our proposed \textit{multiview registration} algorithm as a deep neural network (Fig.~\ref{fig:network_architecture}). In this section, we first describe the architectures used to aproximate $\phi(\cdot), \psi_{\text{init}}(\cdot), \psi_{\text{iter}}(\cdot), \zeta_{\text{init}}(\cdot) $ and $\zeta_{\text{iter}}(\cdot) $, before integrating them into one fully differentiable, end-to-end trainable algorithm.

\vspace{-3mm}
\paragraph{Learned correspondence function}
Our approximation \rev{of} the correspondence function $\phi(\cdot)$ extends a recently proposed fully convolutional 3D feature descriptor FCGF~\cite{Choy2019FCGF} with a soft assignment layer. FCGF operates on sparse tensors~\cite{choy2019Minkowski} and computes $32$ dimensional descriptors for each point of the sparse point cloud in a single pass. Note that the function $\phi(\cdot)$ could be approximated with any of the recently proposed learned feature descriptors~\cite{khoury2017CGF, deng2018ppffold, deng2018ppfnet, gojcic20193DSmoothNet}, but we choose FCGF due to its high accuracy and low computational complexity. 

Let $\mathbf{F}_\mathbf{P}$ and $\mathbf{F}_\mathbf{Q}$ denote the FCGF embeddings of point clouds $\mathbf{P}$ and $\mathbf{Q}$ \rev{obtained using the same network weights}, respectively. Pointwise correspondences $\{\phi(\cdot)\}$ can then be established by a nearest neighbor (NN) search in this high dimensional feature space. However, the selection rule of such hard assignments is not differentiable. We therefore form the NN-selection rule in a probabilistic manner by computing a probability vector $\mathbf{s}$ of the categorical distribution~\cite{plotz2018NN3}. The stochastic correspondence of the point $\mathbf{p}$ in the point cloud $\mathbf{Q}$ is then defined as
\begin{equation}
   \phi(\mathbf{p},\mathbf{Q}) := \mathbf{s}^T\mathbf{Q}, \quad     (\mathbf{s})_{l} := \frac{\exp(-d_l \slash t)}{\sum_{l=1}^{N_\mathbf{Q}} \exp(-d_l \slash t)}
\end{equation}
where $d_l := ||\mathbf{f}_\mathbf{p} - (\mathbf{F}_\mathbf{Q})_l ||_2$, $\mathbf{f}_\mathbf{p}$ is the FCGF embedding of the point $\mathbf{p}$ and $t$ denotes the temperature parameter. In the limit $t \to 0$ the $\phi(\mathbf{p},\mathbf{Q})$ converges to the deterministic NN-search~\cite{plotz2018NN3}. 

We follow~\cite{Choy2019FCGF} and supervise the learning of $\phi(\cdot)$ with a correspondence loss $\mathcal{L}_c$, which is defined as the hardest contrastive loss and operates on the FCGF embeddings
\begin{align*}
    \mathcal{L}_c = \frac{1}{N_\text{FCGF}}&\sum_{(i,j) \in \mathcal{P}} \bigg\{ \big[ d(\mathbf{f}_i,\mathbf{f}_j) - m_p \big]^2_+ \slash |\mathcal{P}|\\ & + 0.5 \big[m_n - \min_{k \in \mathcal{N}}d(\mathbf{f}_i,\mathbf{f}_k)\big]^2_+ \slash |\mathcal{N}_i|\\ & + 0.5\big[m_n - \min_{k \in \mathcal{N}}d(\mathbf{f}_j,\mathbf{f}_k)\big]^2_+ \slash |\mathcal{N}_j|
    \bigg\}
\end{align*}
where $\mathcal{P}$ is a set of all the positive pairs in a FCGF mini batch $N_\text{FCGF}$ and $\mathcal{N}$ is a random subset of all features that is used for the hardest negative mining. $m_p = 0.1$ and $m_n = 1.4$ are the margins for positive and negative pairs respectively. The detailed network architecture of $\phi(\cdot)$ as well as the training configuration and parameters are available in the supp.~material.
\vspace{-3mm}
\paragraph{Deep pairwise registration}
Despite the good performance of the FCGF descriptor, several putative correspondences $\bm{\Gamma}' \subset \bm{\Gamma}$ will be false. Furthermore, the distribution of inliers and outliers does not resemble noise but rather shows regularity~\cite{ranftl2018deepFundamentalMatrix}. We thus aim to learn this regularity from the data using a deep neural network. Recently, several networks representing a complex weighting function for filtering of 2D~\cite{moo2018learn2find,ranftl2018deepFundamentalMatrix, zhang2019oanet} or 3D~\cite{gojcic2019robust} feature correspondences have been proposed. 

Herein, we propose extending the 3D outlier filtering network~\cite{gojcic2019robust} that is based on~\cite{moo2018learn2find} with the order-aware blocks proposed in~\cite{zhang2019oanet}. Specifically, we create a pairwise registration block 
$f_\theta:\mathbb{R}^{N_\mathbf{P}\times 6}\mapsto \mathbb{R}^{N_\mathbf{P}}$
that takes the coordinates of the putative correspondences $\bm{\Gamma}$ as input and outputs weights $\mathbf{w} := \psi_{\text{init}}(\bm{\Gamma}) := \text{tanh}(\text{ReLU}(f_\theta(\bm{\Gamma})))$ that are fed, along with $\bm{\Gamma}$, into the closed form solution of Eq.~\ref{eq:weighted_pairwise} to obtain $\hat{\mathbf{R}}_{ij}$ and $\hat{\mathbf{t}}_{ij}$. Motivated by the results in~\cite{ranftl2018deepFundamentalMatrix, zhang2019oanet} we add another registration block $\psi_{\text{iter}}(\cdot)$ to our network and append the weights $\mathbf{w}$ and the pointwise residuals $\mathbf{r}$ to the original input s.t. $\mathbf{w}^{(k)} := \psi_{iter}(\operatorname{cat}([\bm{\Gamma}^{(k)}, \mathbf{w}^{(k-1)}, \mathbf{r}^{(k-1)}]))$ (see Sec.~\ref{sec:iterative_refinment}). The weights $\mathbf{w}^{(k)}$ are then, again fed together with the initial correspondences $\bm{\Gamma}$ to the closed form solution of Eq.~\ref{eq:weighted_pairwise} to obtain the refined pairwise transformation parameters.  
In order to ensure permutation-invariance of $f_\theta(\cdot)$ a PointNet-like~\cite{qi2017pointnet} architecture that operates on individual correspondences is used in both registration blocks. 
As each branch only operates on individual correspondences, the local 3D context information is gathered in the intermediate layers using symmetric context normalization~\cite{yi2016learning} and order-aware filtering layers~\cite{zhang2019oanet}. The detailed architecture of the registration block is available in the supp.~material. Training of the registration network is supervised using the registration loss $\mathcal{L}_{\text{reg}}$ defined for a batch with $N_\text{reg}$ examples as
\begin{equation}
\mathcal{L}_{\text{reg}} = \alpha_\text{reg}L_{\text{class}} + \beta_\text{reg}L_{\text{trans}}
\end{equation}
loss, where $\mathcal{L}_{\text{class}}$ denotes the binary cross entropy loss and
\begin{equation}
\mathcal{L}_{\text{trans}} = \frac{1}{N_\text{reg}} \sum_{(i,j)}\frac{1}{N_\mathbf{P}}\sum_{l=1}^{N_\mathbf{P}}||\hat{\mathbf{M}}_{ij} \otimes \mathbf{p}_l - \mathbf{M}^{\text{GT}}_{ij} \otimes  \mathbf{p}_l ||_2
\end{equation}
is used to penalize the deviation from the ground truth transformation parameters $\mathbf{M}^{\text{GT}}_{ij}$\rev{. } $\alpha_\text{reg}$ and $\beta_\text{reg}$ are used to control the contribution of the individual loss functions.
\vspace{-3mm}
\paragraph{Confidence estimation block}
Along with the estimated relative transformation parameters $\hat{\mathbf{M}_{ij}}$, the edges of the graph $\mathcal{G}$ encode the confidence $c_{ij}$ in those estimates. Confidence encoded in each edge of the graph consist of (i) the local confidence $c_{ij}^{\text{local}}$ of the pairwise transformation estimation and (ii) the global confidence $c_{ij}^{\text{global}}$ derived from the transformation synchronization. We formulate the estimation of $c_{ij}^{\text{local}}$ as a classification task and argue that some of the required information is encompassed in the features of the second-to-last layer of the registration block.
Let $\mathbf{X}^{\text{conf}}_{ij} = f_\theta^{(-2)}(\cdot)$ denote the output of the second-to-last layer of the registration block, we propose an \textit{overlap pooling layer} $f_\text{overlap}$ that extracts a global feature $\mathbf{x}^\text{conf}_{ij}$ by performing the weighted average pooling as
\begin{equation}
    \mathbf{x}^\text{conf}_{ij} = \mathbf{w}_{ij}^{{\text{T}}}\mathbf{X}_{ij}^{\text{conf}}.
\end{equation}
The obtained global feature is concatenated with the ratio of inliers $\delta_{ij}$ (i.e., the number of correspondences whose weights are higher than a given threshold) and fed to the confidence estimation network with three fully connected layers $(129-64-32-1)$, followed by a ReLU activation function. The local confidence can thus be expressed as
\begin{equation}
    c_{ij}^{\mathrm{local}} := \zeta_{\text{init}}(\bm{\Gamma}) := \operatorname{MLP}(\operatorname{cat}([\mathbf{x}^\text{conf}_{ij}, \delta_{ij}]))
    \label{eq:local_conf}
\end{equation}
The training of the confidence estimation block is supervised with the confidence loss function $\mathcal{L}_{\text{conf}} =\frac{1}{N} \sum_{(i,j)} \text{BCE}(c_{ij}^{\mathrm{local}},c^{\text{GT}}_{ij})$ ($N$ denotes the number of cloud pairs), where BCE refers to the binary cross entropy and the ground truth confidence $c^{\text{GT}}_{ij}$ labels are computed on the fly by thresholding the angular error $\tau_a = \operatorname{arccos}{(\frac{\text{Tr}(\hat{\mathbf{R}}^T_{ij}\mathbf{R}^\text{GT}_{ij})-1}{2})}$.

The $\zeta_{\text{init}}(\cdot)$ incorporates the \textit{local} confidence in the relative transformation parameters. On the other hand, the output of the transformation synchronization layer provides the information how the input relative transformations agree globally with the other edges. In fact, traditional synchronization algorithms~\cite{chatterjee2013efficient, arrigoni2016se3sync, huang2017transSync} only use this \textit{global} information to perform the reweighting of the edges in the iterative solutions, because they do not have access to the \textit{local} confidence information. \textit{Global} confidence in the relative transformation parameters $c_{ij}^{\text{global}}$ can be expressed with the Cauchy weighting function~\cite{holland1977cauchy, arrigoni2016se3sync}
\begin{equation}
\label{eq:glob_w}
    c^{\text{global}}_{ij} = {1}\slash{(1+ r^*_{ij}\slash b)}
\end{equation}
where $r^*_{ij} = ||\hat{\mathbf{M}}_{ij} - \mathbf{M}_i^*\mathbf{M}_j^{*^{T}}||_F$ and following~\cite{holland1977cauchy,arrigoni2016se3sync} $b=1.482 \mkern3mu \gamma \mkern3mu \operatorname{med}(|\mathbf{r}^*-\operatorname{med}(\mathbf{r}^*)|)$ with $\operatorname{med}(\cdot)$ denoting the median operator and $\mathbf{r}^*$ the vectorization of residuals $r^*_{ij}$. 
Since \textit{local} and \textit{global} confidence provide complementary information about the relative transformation parameters, we combine them into a joined confidence $c_{ij}$ using their harmonic mean:
\begin{equation}
    \label{eq:hm_f}
    c_{ij} := \zeta_{iter}(c^\text{local}_{ij},c^\text{global}_{ij}) := \frac{(1+\beta^2)c^{\text{global}}_{ij}\cdot c^{\text{local}}_{ij}}{\beta^2 c^{\text{global}}_{ij} + c^{\text{local}}_{ij}}
\end{equation}
where the $\beta$ balances the contribution of the local and global confidence estimates and is learned during training. 
% I have updated all the values such that they agree with the FCGF (They are different from PPFNet)
\begin{table}[!t]
	\centering
	\resizebox{\linewidth}{!}{\begin{tabular}{l|cccccccccccccccc}
			\hline
			& \textit{3DMatch} & \textit{CGF} & \textit{PPFNet} & \textit{3DR} & \textit{3DSN} & \textit{FCGF} & \multicolumn{2}{c}{\textit{Ours}} \\
			& \cite{zeng20163dmatch} & \cite{khoury2017CGF} & \cite{deng2018ppfnet}& \cite{deng20193d} & \cite{gojcic20193DSmoothNet} & \cite{Choy2019FCGF} & \textit{1-iter} & \textit{4-iter} \\
			\hline\hline
            Kitchen &     $0.85$ & $0.72$ & $0.90$ & $0.80$ & $\mathbf{0.96}$ & $0.95$ & $\mathbf{0.96}$ & $0.98$ \\
            Home 1  &     $0.78$ & $0.69$ & $0.58$ & $0.81$ & $0.88$ & $0.91$ & $\mathbf{0.92}$ & $0.93$ \\
            Home 2  &     $0.61$ & $0.46$ & $0.57$ & $0.70$ & $\mathbf{0.79}$ & $0.72$ & $0.70$ & $0.73$ \\
            Hotel 1 &     $0.79$ & $0.55$ & $0.75$ & $0.73$ & $\mathbf{0.95}$ & $0.93$ & $\mathbf{0.95}$ & $0.97$ \\
            Hotel 2 &     $0.59$ & $0.49$ & $0.68$ & $0.67$ & $0.83$ & $0.88$ & $\mathbf{0.90}$ & $0.90$ \\
            Hotel 3 &     $0.58$ & $0.65$ & $0.88$ & $\mathbf{0.94}$ & $0.92$ & $0.81$ & $0.89$ & $0.89$ \\
            Study   &     $0.63$ & $0.48$ & $0.68$ & $0.70$ & $0.84$ & $\mathbf{0.86}$ & $\mathbf{0.86}$ & $0.92$ \\
            MIT Lab &     $0.51$ & $0.42$ & $0.62$ & $0.62$ & $0.76$ & $\mathbf{0.82}$ & $0.78$ & $0.78$ \\
			\hline
			Average &     $0.67$ & $0.56$ & $0.71$ & $0.75$ & $0.86$ & $0.86$ & $\mathbf{0.87}$ & $0.89$\\
			\hline
	\end{tabular}}
	\caption{Registration recall on 3DMatch data set. \textit{1-iter} and \textit{4-iter} denote the result of the pairwise registration network and input to the $4$th Trasnf-Sync laser, respectively. Best results, except for \textit{4-iter} that is informed by the global information, are shown in bold. }
	\label{tab:reg_recall3DMatch_dataset}\vspace{-3mm}
\end{table}
\vspace{-3mm}
\paragraph{End-to-end multiview 3D registration}
The individual parts of the network are connected into an end-to-end multiview 3D registration algorithm as shown in
Fig.~\ref{fig:network_architecture}\footnote{The network is implemented in Pytorch~\cite{paszke2017automatic}. A pseudo-code of the proposed approach is provided in the supp.~material.}. We pre-train the individual sub-networks (training details available in the supp.~material) before fine-tuning the whole model in an end-to-end manner on the 3DMatch data set~\cite{zeng20163dmatch} using the official train/test data split. In fine-tuning we use $N_{\text{FCGF}} = 4$ to extract the FCGF features and randomly sample feature vectors of 2048 points per fragment. These features are used in the soft assignment (softNN) to form the putative correspondences of $N_\mathcal{S} \choose 2$ point clouds pairs\footnote{We assume a fully connected graph during training but are able to consider the connectivity information, if provided.}, which are fed to the pairwise registration network. The output of the pairwise registration is used to build the graph, which is input to the transformation synchronization layer. The iterative refinement of the transformation parameters is performed four times. We supervise the fine tuning using the joint multiview registration loss 
\begin{equation}
    \mathcal{L} = \mathcal{L}_\text{c} + \mathcal{L}_\text{reg} + \mathcal{L}_\text{conf} + \mathcal{L}_\text{sync}
\end{equation}
where the transformation synchronization $\mathcal{L}_{sync}$ loss reads
\begin{equation}
    \label{eq:sync_rot_loss}
    \mathcal{L}_{sync} = \frac{1}{N}\sum_{(i,j)}(\|\mathbf{R}^*_{ij} - \mathbf{R}^{GT}_{ij}\|_F + \|\mathbf{t}^*_{ij} - \mathbf{t}^{GT}_{ij}\|_2).
\end{equation}
We fine-tune the whole network for $2400$ iterations using Adam optimizer~\cite{Kingma2015ADAM} with a learning rate of $5 \times 10^{-6}$.

\begin{table}[!t]
    \setlength{\tabcolsep}{4pt}
	\centering
    \resizebox{0.8\linewidth}{!}{
    \begin{tabular}{l|cc|c}
            \hline
            &\multicolumn{2}{c|}{Per fragment pair}& Whole scene\\
			\hline
			& NN search & Model estimation  & Total time  \\
			& [s] &  [s]& [s]  \\
			\hline
			\hline
			RANSAC           &$0.38$&$0.23$& $1106.3$\\
		    Ours (softNN)    &$0.10$&$0.01$&  $80.3$ \\
			\hline
	\end{tabular}}
	\caption{Average run-time for estimating the pairwise transformation parameters of one fragment pair on \textit{3DMatch} dataset.\rev{Note, the GPU implementation of the soft assignments is faster than the CPU based kd-tree NN search.}}
	\label{tab:timing_3DMatch}
\end{table}
\section{Experiments}
We conduct the evaluation of our approach on the  publicly available benchmark datasets \textit{3DMatch}~\cite{zeng20163dmatch}, \textit{Redwood}~\cite{choi2015robust} and \textit{ScanNet}~\cite{dai2017scannet}. First, we evaluate the performance, efficiency, and the generalization capacity of the proposed pairwise registration algorithm  on \textit{3DMatch} and \textit{Redwood} dataset respectively (Sec. \ref{subsec:pw_3dmatch}). We then evaluate the whole pipeline on the global registration of the point cloud fragments generated from RGB-D images, which are part of the \textit{ScanNet} dataset~\cite{dai2017scannet}.
\begin{table*}[!t]
	\centering
	\setlength{\tabcolsep}{7pt}
    \resizebox{\linewidth}{!}{\begin{tabular}{cl|cccccl|cccccl}
    \hline
         &Methods & \multicolumn{6}{c|}{Rotation Error} &\multicolumn{6}{c}{Translation Error ($\mathrm{m}$)}  \\
         \hline
         & & $3^\circ$&$5^\circ$&$10^\circ$&$30^\circ$&$45^\circ$&Mean/Med.&0.05&0.1&0.25&0.5&0.75&Mean/Med.\\
         \hline\hline
         \multirow{2}{*}{\shortstack{Pairwise\\(All)}}&\textit{FGR}~\cite{zhou2016FGR} &9.9 & 16.8&23.5 &31.9 &38.4 &$76.3^\circ$/-& 5.5& 13.3& 22.0&29.0 &36.3 & 1.67/- \\
         &\textit{Ours} ($1^{st}$ iter.)&32.6 & 37.2 & 41.0 & 46.5 & 49.4& $65.9^\circ$/$48.8^\circ$&25.1 & 34.1&40.0 &43.4 &46.8 &1.37/0.94 \\
         \hline\hline
         \multirow{2}{*}{\shortstack{Edge Pruning\\(All)}}&\textit{Ours} ($4^{th}$ iter.) &34.3 & 38.7 & 42.2 & 48.2 & 51.9& $62.3^\circ$/$37.0^\circ$&26.7 & 35.7&41.8 &45.5 &49.4 &1.26/0.78 \\
         &\textit{Ours} (After Sync.) &40.7 & 45.7 & 50.8 & 56.2 & 58.4& $52.2^\circ$/$9.0^\circ$&29.3 & 42.1 & 50.9 & 54.7 & 58.3 & 0.96/0.20 \\
         \hline\hline
         \multirow{5}{*}{\shortstack{FGR\\(Good)}}&\textit{FastGR}~\cite{zhou2016FGR}&12.4&21.4&29.5&38.6&45.1&$68.8^\circ$/-&7.7&17.6&28.2&36.2&43.4&1.43/-\\
         &\textit{GeoReg} (\textit{FGR})~\cite{choi2015robust}&0.2&0.6&2.8&16.4&27.1&$87.2^\circ$/-&0.1&0.7&4.8&16.4&28.4&1.80/-\\
         &\textit{EIGSE3} (\textit{FGR})~\cite{arrigoni2016se3sync}&1.5&4.3&12.1&34.5&47.7&$68.1^\circ$/-&1.2&4.1&14.7&32.6&46.0&1.29/-\\
         &\textit{RotAvg} (\textit{FGR}~\cite{chatterjee2018robust})&6.0&10.4&17.3&36.1&46.1&$64.4^\circ$/-&3.7&9.2&19.5&34.0&45.6&1.26/-\\
         &\textit{L2Sync} (\textit{FGR})~\cite{huang2019learningTransSync}&34.4&41.1&49.0&58.9&62.3&$42.9^\circ$/-&2.0&7.3&22.3&36.9&48.1&1.16/-\\
         \hline\hline 
         \multirow{3}{*}{\shortstack{Ours\\(Good)}}&\textit{EIGSE3}~\cite{arrigoni2016se3sync}& 63.3& 70.2& 75.6&80.5&81.6 &$23.0^\circ$/$1.7^\circ$ & 42.2&58.5 & 69.8&76.9& 79.7&0.45/0.06 \\
         &\textit{Ours} ($1^{st}$ iter.) & 57.7& 65.5& 71.3&76.5&78.1 &$28.3^\circ$/$1.9^\circ$ & 44.8&60.3 & 69.6&73.1& 75.5&0.57/0.06 \\
         &\textit{Ours} ($4^{th}$ iter.) & 60.6& 68.3& 73.7&78.9&81.0 &$24.2^\circ$/$1.8^\circ$  & 47.1&63.3 & 72.2&76.2 &78.7 &0.50/\textbf{0.05} \\
         &\textit{Ours} (After Sync) &\textbf{65.8} & \textbf{72.8} & \textbf{77.6} & \textbf{81.9} & \textbf{83.2}& $\textbf{20.3}^\circ$/$\textbf{1.6}^\circ$&\textbf{48.4} & \textbf{67.2}&\textbf{76.5} &\textbf{79.7} &\textbf{82.0} &\textbf{0.42}/\textbf{0.05} \\
         \hline
    \end{tabular}}
    \vspace{.1ex}
    \caption{Multiview registration evaluation on \textit{ScanNet}~\cite{dai2017scannet} dataset. We report the ECDF values for rotation and translation errors. Best results are shown in bold.}%
    \label{tab:cmp_l2s_scannet}\vspace{-3mm}
\end{table*}

\subsection{Pairwise registration performance}\label{subsec:pw_3dmatch}
We begin by evaluating the pairwise registration part of our algorithm on a traditional geometric registration task. We compare the results of our method to the state-of-the-art data-driven feature descriptors 3DMatch~\cite{zeng20163dmatch}, CGF~\cite{khoury2017CGF}, PPFNet~\cite{deng2018ppfnet}, 3DSmoothNet (3DS)~\cite{gojcic20193DSmoothNet}, and FCGF~\cite{Choy2019FCGF}, which is also used as part of our algorithm\rev{, as well as to a recent network based registration algorithm 3DR~\cite{deng20193d}.} Following the evaluation procedure of \textit{3DMatch}~\cite{zeng20163dmatch} we complement all the \rev{descriptor based methods} with the RANSAC-based transformation parameter estimation. For our approach we report the results after the pairwise registration network (1-iter in Tab.~\ref{tab:reg_recall3DMatch_dataset}) as well as the the output of the $\psi_{\text{iter}} (\cdot)$ in the $4^\text{th}$ iteration (4-iter in Tab.~\ref{tab:reg_recall3DMatch_dataset}). The latter is already informed with the global information and serves \rev{primarily} as verification that with the iterations our input to the \textit{Transf-Sync} layer improves. Consistent with the \textit{3DMatch} evaluation procedure, we report the average recall per scene as well as for the whole dataset in Tab.~\ref{tab:reg_recall3DMatch_dataset}. 

The registration results show that our approach reaches the highest recall among all the evaluated methods. More importantly, it indicates that using the same features (FCGF), our method can outperform RANSAC-based estimation of the transformation parameters, while having a much lower time complexity (Tab.~\ref{tab:timing_3DMatch}). The comparison of the results of 1-iter and 4-iter also confirms the intuition that feeding the residuals and weights of the previous estimation back to the pairwise registration block helps refining the estimated pairwise transformation parameters.

\begin{figure}
	\begin{center}
		\includegraphics[width=\columnwidth]{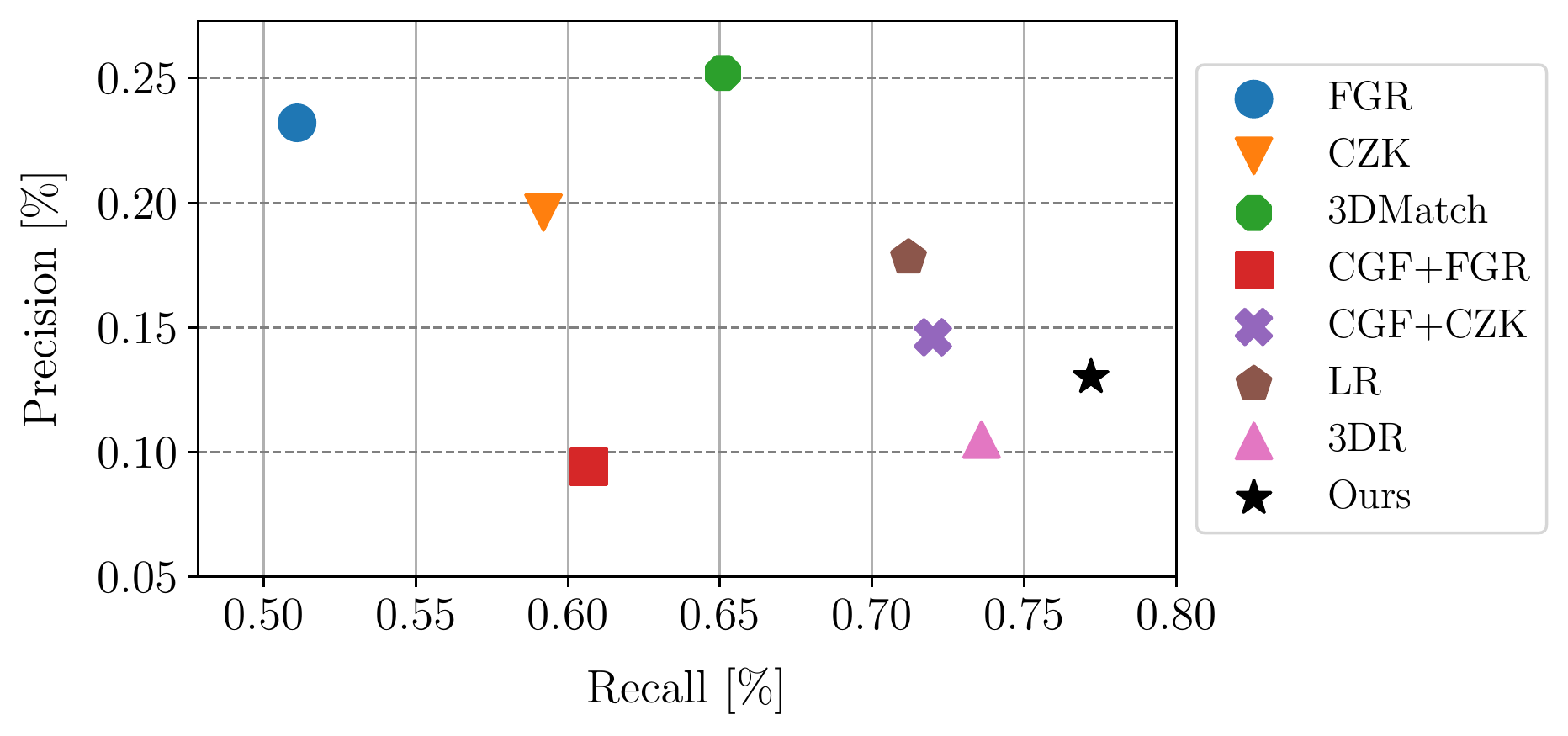}
	\end{center}
	\vspace{-3.5ex}
	\caption{Registration results on the \textit{Redwood indoor} dataset.}
	\label{fig:redwood_indoor}\vspace{-3mm}
\end{figure}

\vspace{2mm}\noindent\textbf{Generalization to other domains\, }
In order to test if our pairwise registration model can generalize to new datasets and unseen domains, we perform a generalization evaluation on a synthetic indoor dataset \textit{Redwood indoor}~\cite{choi2015robust}. We follow the evaluation protocol of~\cite{choi2015robust} and report the average registration recall and precision across all four scenes. We compare our approach to the recent data driven approaches 3DMatch~\cite{zeng20163dmatch}, CGF~\cite{khoury2017CGF}+FGR~\cite{zhou2016FGR} or \rev{CZK~\cite{choi2015robust}}, RelativeNet (RN)~\cite{deng20193d}, \rev{3DR~\cite{deng20193d}} and traditional methods CZK~\cite{choi2015robust} and Latent RANSAC (LR)~\cite{korman2018latent}.
Fig.~\ref{fig:redwood_indoor} shows that our approach can achieve $\approx 4$ percentage points higher recall than state-of-the-art without being trained on synthetic data and thus confirming the good generalization capacity of our approach. Note that while the average precision across the scenes is low for all the methods, several works~\cite{choi2015robust, khoury2017CGF, deng20193d} show that the precision can easily be increased using pruning without almost any loss in the recall.

\subsection{Multiview registration performance} We \rev{finally} evaluate the performance of our \rev{complete} method on the task of multiview registration using the \textit{ScanNet}~\cite{dai2017scannet} dataset. \textit{ScanNet} is a large RGBD dataset of indoor scenes. It provides the reconstructions, ground truth camera poses and semantic segmentations for $1513$ scenes. To ensure a fair comparison, we follow~\cite{huang2019learningTransSync} and use the same $32$ randomly sampled scenes for evaluation. For each scene we randomly sample $30$ RGBD images that are 20 frames apart and convert them to point clouds. The temporal sequence of the frames is discarded. In combination with the large \rev{temporal} gap between the frames, this makes the test setting extremely challenging. Different to~\cite{huang2019learningTransSync}, we do not train our network on \textit{ScanNet}, but rather perform direct generalization of the network trained on the 3DMatch dataset. 

\vspace{2mm}\noindent\textbf{Evaluation protocol\, }
We use the standard evaluation protocol~\cite{chatterjee2013efficient, huang2019learningTransSync} and report the empirical cumulative distribution function (ECDF) for the angular $a_e$ and translation $t_e$ deviations defined as  
\begin{equation}
\resizebox{0.85\linewidth}{!}{$
     a_e = \operatorname{arccos}(\frac{\operatorname{Tr}(\mathbf{R}_{ij}^{\mathrm{*}^{\mathrm{T}}}\mathbf{R}_{ij}^{\mathrm{GT}}) - 1}{2}) \quad t_e = \|\mathbf{t}_{ij}^{\mathrm{GT}} - \mathbf{t}_{ij}^{\mathrm{*}}\|_2$}
\label{eq:ae_te}
\end{equation}

The ground truth rotations $\mathbf{R^{\text{GT}}}$ and translations $\mathbf{t^{\text{GT}}}$ are provided by the authors of \textit{ScanNet}~\cite{dai2017scannet}. In Tab.~\ref{tab:cmp_l2s_scannet} we report the results for three different scenarios. "FGR~(Good)" and "Ours~(Good)" denote the scenarios in which we follow~\cite{huang2019learningTransSync} and use the computed pairwise registrations to prune the edges before the transformation synchronization if the median point distance in the overlapping\footnote{The overlapping regions are defined as parts, where after transformation, the points are less than $0.2$m away from the other point cloud.~\cite{huang2019learningTransSync}} region after the transformation is larger than $0.1$m (FGR) or $0.05$m (ours). The EIGSE3 in "Ours (Good)" is initialized using our pairwise estimates. On the other hand, "all" denotes the scenario in which all $N_{\mathcal{S}} \choose 2$ pairs are used to build the graph. In all scenarios we prune the edges of the graph if the confidence estimation in the relative transformation parameters of that edge $c^{\text{local}}_{ij}$ drops below $\tau_p=0.85$. This threshold was determined on \textit{3DMatch} dataset and its effect on the performance of our approach is analyzed in detail in the supp.~material. \rev{If during the iterations the pruning of the edges yields a disconnected graph we simply report the last valid values for each node before the graph becomes disconnected. A more sophisticated handling of the edge pruning and disconnected graphs is left for future work.}
%we continue the computation on the largest subgraph and  report the last valid values for each node that was disconnected.}
%CZ: actually there may be a bit misunderstanding. Continuing on the largest subgraph is what have been used in Tolga's paper. In our paper we just report the last valid values for each node before the graph gets disconnected.

\vspace{2mm}\noindent\textbf{Analysis of the results}
As shown in Tab.~\ref{tab:cmp_l2s_scannet} our approach can achieve a large improvement on the multiview registration tasks when compared to the baselines. Not only are the initial pairwise relative transformation parameters estimated using our approach more accurate than the ones of FGR~\cite{zhou2016FGR}, but they can also be further improved in the subsequent iterations. This clearly confirms the benefit of the feed-back loop of our algorithm. 
Furthermore even when directly considering all input edges our approach still proves dominant, even when considering the results of the scenario "Good" for our competitors. More qualitative results of the multiview registration evaluation\rev{, including the failure cases,} are available in the supp.~material.

\vspace{3mm}\noindent\textbf{Computational complexity}
Low computational costs of pairwise and multiview registration is important for various fields like augmented reality or robotics. We first compare computation time of our pairwise registration component to RANSAC. In Tab.~\ref{tab:timing_3DMatch} we report the average time needed to register one fragment pair of the \textit{3DMatch} dataset as well as one whole scene. All timings were performed on a standalone computer with Intel(R) Core(TM) i7-7700K CPU @ 4.20GHz, GeForce GTX 1080, and 32 GB RAM. Average time of performing softNN for a fragment pair is about $0.1\text{s}$, which is a approxiately four times faster than traditional nearest neighbor search (implemented using scikit-learn~\cite{pedregos2011scikit}). An even larger speedup (about $23$ times) is gained in the model estimation stage, where our approach requires a single forward pass (constant time) compared to up to 50000 iterations of RANSAC when the inlier ratio is $5\%$ and the desired confidence $0.99$\footnote{\rev{We use the CPU-based RANSAC implementation that is provided in the original evaluation code of 3DMatch dataset~\cite{zeng20163dmatch}.}}.
This results in an overall run-time of about $80\text{s}$ for our entire multiview approach (including the feature extraction and transformation synchronization) for the \textit{Kitchen} scene with 1770 fragment pairs. In contrast, feature extraction and pairwise estimation of transformation parameters with RANSAC takes $>1100$s. This clearly shows the efficiency of our method, being $>13$ times faster to compute (for a scene with $60$ fragments).

\subsection{Ablation study}
\begin{figure}
	\begin{center}
		\includegraphics[width=\columnwidth]{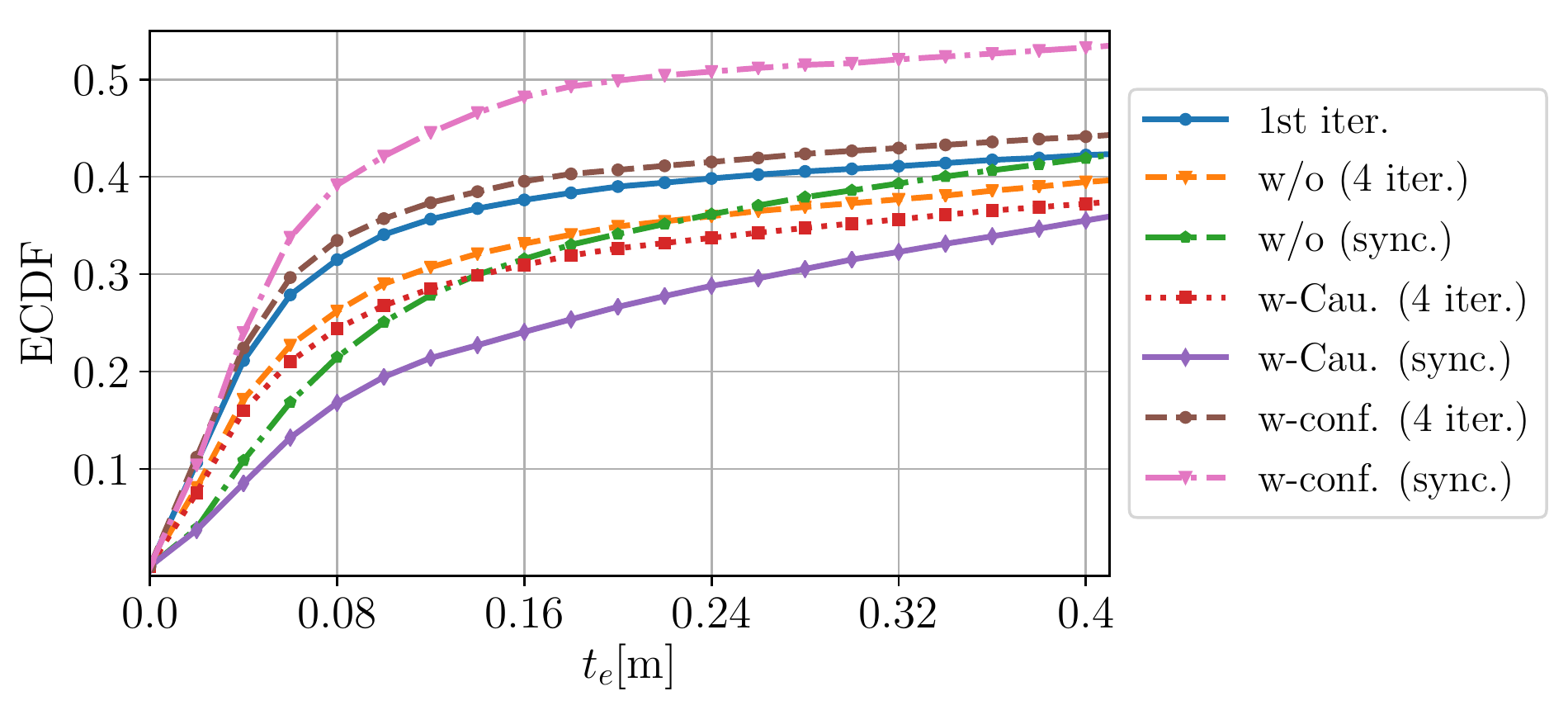}
	\end{center}
	\vspace{-3.5ex}
	\caption{Ablation study on the \textit{ScanNet} dataset.}
	\label{fig:abl_stu_scannet}\vspace{-3mm}
\end{figure}

To get a better intuition how much the individual novelties in our approach contribute to the final performance, we carry out an ablation study on the \textit{ScanNet}~\cite{dai2017scannet} dataset. In particular, we analyze the proposed edge pruning scheme based on the confidence estimation block and \emph{Cauchy} function as well as the impact of the iterative refinement of the relative transformation parameters.\footnote{Additional results of the ablation study are included in the supp.~material.} The results of the ablation study are presented in Fig.~\ref{fig:abl_stu_scannet}.

\paragraph{Benefit from the iterative refinement}
We motivate the iterative refinement of the transformation parameters that are input to the \textit{Transf-Sync} layer with a notion that the weights and residuals provide additional ques for their estimation. Results in Fig.~\ref{fig:abl_stu_scannet} confirm this assumption. The input relative parameters in the 4-th iteration are approximately $2$ percentage points better that the initial estimate. On the other hand, Fig.~\ref{fig:abl_stu_scannet} shows that at the high presence of outliers or inefficient edge pruning (see e.g., the results w/o edge pruning) the weights and the residuals actually provide a negative bias and worsen the results.

\vspace{3mm}\noindent\textbf{Edge pruning scheme}
There are several possible ways to implement the pruning of the presumable outlier edges. In our experiments we prune the edges based on the output of the confidence estimation block (w-conf.). Other options are to realize this step using the global confidence, i.e. the Cauchy weights defined in \eqref{eq:glob_w}~(w-Cau.) or not performing this at all (w/o).
Fig.~\ref{fig:abl_stu_scannet} clearly shows the advantage of using our confidence estimation block (gain of more than $20$ percentage points). Even more, due to preserving a large amount of outliers, alternative approaches preform even worse than the pairwise registration.

\section{Conclusions}\label{sec:conclusions}
We have introduced an end-to-end learnable, multiview point cloud registration algorithm. Our method departs from the common two-stage approach and directly learns to register all views in a globally consistent manner. We augment the 3D descriptor FCGF~\cite{Choy2019FCGF} by a soft correspondence layer that pairs all the scans to compute initial matches, which are fed to a differentiable pairwise registration block resulting in transformation parameters as well as weights. A pose graph is constructed and a novel, differentiable iterative transformation synchronization layer globally refines weights and transformations.
Experimental evaluation on common benchmark datasets show that our method outperforms state-of-the-art by more than 25 percentage points on average regarding the rotation error statistics. Moreover, our approach is $>13$ times faster than RANSAC-based methods in a multiview setting of 60 scans, and generalizes better to new scenes ($\approx4$ percentage points higher recall on Redwood indoor compared to state-of-the-art).

\vspace{2mm}\noindent\footnotesize\textbf{Acknowledgements}. This work is partially supported by Stanford-Ford Alliance, NSF grant IIS-1763268, Vannevar Bush Faculty Fellowship, Samsung GRO program and the Stanford SAIL Toyota Research Center. We thank NVIDIA Corp.~for providing the GPUs used in this work.

%{\small
%\bibliographystyle{ieee_fullname}
%\bibliography{cvpr2020}

{\small
\begin{thebibliography}{10}\itemsep=-1pt

\bibitem{aiger20084PCS}
Dror Aiger, Niloy~J Mitra, and Daniel Cohen-Or.
\newblock 4-points congruent sets for robust pairwise surface registration.
\newblock In {\em ACM transactions on graphics (TOG)}, number~3, 2008.

\bibitem{arie2012global}
Mica Arie-Nachimson, Shahar~Z Kovalsky, Ira Kemelmacher-Shlizerman, Amit
  Singer, and Ronen Basri.
\newblock Global motion estimation from point matches.
\newblock In {\em International Conference on 3D Imaging, Modeling, Processing,
  Visualization \& Transmission}, 2012.

\bibitem{arrigoni2014robust}
Federica Arrigoni, Luca Magri, Beatrice Rossi, Pasqualina Fragneto, and Andrea
  Fusiello.
\newblock Robust absolute rotation estimation via low-rank and sparse matrix
  decomposition.
\newblock In {\em IEEE International Conference on 3D Vision (3DV)}, pages
  491--498, 2014.

\bibitem{arrigoni2016se3sync}
Federica Arrigoni, Beatrice Rossi, and Andrea Fusiello.
\newblock Spectral synchronization of multiple views in se(3).
\newblock {\em SIAM Journal on Imaging Sciences}, 9(4):1963–1990, 2016.

\bibitem{bernard2015MAtranssync}
Florian Bernard, Johan Thunberg, Peter Gemmar, Frank Hertel, Andreas Husch, and
  Jorge Goncalves.
\newblock A solution for multi-alignment by transformation synchronisation.
\newblock In {\em IEEE Conference on Computer Vision and Pattern Recognition
  (CVPR)}, pages 2161--2169, 2015.

\bibitem{besl1992ICP}
PJ Besl and Neil~D McKay.
\newblock A method for registration of 3-d shapes.
\newblock {\em IEEE Transactions on Pattern Analysis and Machine Intelligence
  (TPAMI)}, 14(2):239--256, 1992.

\bibitem{Bhattacharya2019robustReg}
Uttaran Bhattacharya and Venu~Madhav Govindu.
\newblock Efficient and robust registration on the 3d special euclidean group.
\newblock In {\em The IEEE International Conference on Computer Vision (ICCV)},
  2019.

\bibitem{birdal2015point}
Tolga Birdal and Slobodan Ilic.
\newblock Point pair features based object detection and pose estimation
  revisited.
\newblock In {\em IEEE International Conference on 3D Vision (3DV)}, 2015.

\bibitem{birdal2017cad}
Tolga Birdal and Slobodan Ilic.
\newblock Cad priors for accurate and flexible instance reconstruction.
\newblock In {\em IEEE International Conference on Computer Vision (ICCV)},
  2017.

\bibitem{birdal2019probabilistic}
Tolga Birdal and Umut Simsekli.
\newblock Probabilistic permutation synchronization using the riemannian
  structure of the birkhoff polytope.
\newblock In {\em Proceedings of the IEEE Conference on Computer Vision and
  Pattern Recognition}, pages 11105--11116, 2019.

\bibitem{birdal2018bayesian}
Tolga Birdal, Umut Simsekli, Mustafa~Onur Eken, and Slobodan Ilic.
\newblock Bayesian pose graph optimization via bingham distributions and
  tempered geodesic mcmc.
\newblock In {\em Advances in Neural Information Processing Systems (NIPS)},
  pages 308--319, 2018.

\bibitem{chatterjee2018robust}
A Chatterjee and VM Govindu.
\newblock Robust relative rotation averaging.
\newblock {\em IEEE transactions on pattern analysis and machine intelligence},
  40(4):958--972, 2018.

\bibitem{chatterjee2013efficient}
Avishek Chatterjee and Venu Madhav~Govindu.
\newblock Efficient and robust large-scale rotation averaging.
\newblock In {\em Proceedings of the IEEE International Conference on Computer
  Vision}, pages 521--528, 2013.

\bibitem{choi2015robust}
Sungjoon Choi, Qian-Yi Zhou, and Vladlen Koltun.
\newblock Robust reconstruction of indoor scenes.
\newblock In {\em IEEE Conference on Computer Vision and Pattern Recognition
  (CVPR)}, 2015.

\bibitem{choy2019Minkowski}
Christopher Choy, JunYoung Gwak, and Silvio Savarese.
\newblock 4d spatio-temporal convnets: Minkowski convolutional neural networks.
\newblock In {\em IEEE Conference on Computer Vision and Pattern Recognition
  (CVPR)}, pages 3075--3084, 2019.

\bibitem{Choy2019FCGF}
Christopher Choy, Jaesik Park, and Vladlen Koltun.
\newblock Fully convolutional geometric features.
\newblock In {\em The IEEE International Conference on Computer Vision (ICCV)},
  pages 8958--8966, 2019.

\bibitem{cui2015globalsfm}
Zhaopeng Cui and Ping Tan.
\newblock Global structure-from-motion by similarity averaging.
\newblock In {\em Proceedings of the IEEE International Conference on Computer
  Vision}, pages 864--872, 2015.

\bibitem{dai2017scannet}
Angela Dai, Angel~X. Chang, Manolis Savva, Maciej Halber, Thomas Funkhouser,
  and Matthias Nie{\ss}ner.
\newblock Scannet: Richly-annotated 3d reconstructions of indoor scenes.
\newblock In {\em Proc. Computer Vision and Pattern Recognition (CVPR), IEEE},
  2017.

\bibitem{Deng2018PPFFoldNetUL}
Haowen Deng, Tolga Birdal, and Slobodan Ilic.
\newblock Ppf-foldnet: Unsupervised learning of rotation invariant 3d local
  descriptors.
\newblock In {\em European conference on computer vision (ECCV)}, 2018.

\bibitem{deng2018ppffold}
Haowen Deng, Tolga Birdal, and Slobodan Ilic.
\newblock Ppf-foldnet: Unsupervised learning of rotation invariant 3d local
  descriptors.
\newblock In {\em European Conference on Computer Vision (ECCV)}, pages
  602--618, 2018.

\bibitem{deng2018ppfnet}
Haowen Deng, Tolga Birdal, and Slobodan Ilic.
\newblock Ppfnet: Global context aware local features for robust 3d point
  matching.
\newblock In {\em IEEE Conference on Computer Vision and Pattern Recognition
  (CVPR)}, pages 195--205, 2018.

\bibitem{deng20193d}
Haowen Deng, Tolga Birdal, and Slobodan Ilic.
\newblock 3d local features for direct pairwise registration.
\newblock In {\em IEEE Conference on Computer Vision and Pattern Recognition
  (CVPR)}, 2019.

\bibitem{ding2019}
Li Ding and Chen Feng.
\newblock {DeepMapping:} {Unsupervised map} estimation from multiple point
  clouds.
\newblock In {\em IEEE Conference on Computer Vision and Pattern Recognition
  (CVPR)}, pages 8650--8659, 2019.

\bibitem{drost2010model}
Bertram Drost, Markus Ulrich, Nassir Navab, and Slobodan Ilic.
\newblock Model globally, match locally: Efficient and robust 3d object
  recognition.
\newblock In {\em IEEE Conference on Computer Vision and Pattern Recognition
  (CVPR)}, pages 998--1005, 2010.

\bibitem{fantoni2012accurate}
Simone Fantoni, Umberto Castellani, and Andrea Fusiello.
\newblock Accurate and automatic alignment of range surfaces.
\newblock In {\em 2012 Second International Conference on 3D Imaging, Modeling,
  Processing, Visualization \& Transmission}, pages 73--80. IEEE, 2012.

\bibitem{fischer1981RANSAC}
Martin~A. Fischler and Robert~C. Bolles.
\newblock Random sample consensus: A paradigm for model fitting with
  applications to image analysis and automated cartography.
\newblock {\em Commun. ACM}, 24(6):381--395, 1981.

\bibitem{flint2007}
A. Flint, A. Dick, and A. {van den Hangel}.
\newblock {Thrift: Local 3D structure recognition}.
\newblock In {\em 9th Biennial Conference of the Australian Pattern Recognition
  Society on Digital Image Computing Techniques and Applications}, 2007.

\bibitem{gojcic20193DSmoothNet}
Zan Gojcic, Caifa Zhou, Jan~D Wegner, and Andreas Wieser.
\newblock The perfect match: 3d point cloud matching with smoothed densities.
\newblock In {\em IEEE Conference on Computer Vision and Pattern Recognition
  (CVPR)}, 2019.

\bibitem{gojcic2019robust}
Zan Gojcic, Caifa Zhou, and Andreas Wieser.
\newblock Robust pointwise correspondences for point cloud based deformation
  monitoring of natural scenes.
\newblock In {\em 4th Joint International Symposium on Deformation Monitoring
  (JISDM)}, 2019.

\bibitem{govindu2004liesync}
Venu~Madhav Govindu.
\newblock Lie-algebraic averaging for globally consistent motion estimation.
\newblock In {\em IEEE Conference on Computer Vision and Pattern Recognition
  (CVPR)}, pages 684--691, 2004.

\bibitem{gross2019alignnet}
Johannes Gro{\ss}, Aljo{\v{s}}a O{\v{s}}ep, and Bastian Leibe.
\newblock Alignnet-3d: Fast point cloud registration of partially observed
  objects.
\newblock In {\em 2019 International Conference on 3D Vision (3DV)}, pages
  623--632. IEEE, 2019.

\bibitem{he2016deep}
Kaiming He, Xiangyu Zhang, Shaoqing Ren, and Jian Sun.
\newblock Deep residual learning for image recognition.
\newblock In {\em IEEE Conference on Computer Vision and Pattern Recognition
  (CVPR)}, pages 770--778, 2016.

\bibitem{hinterstoisser2016going}
Stefan Hinterstoisser, Vincent Lepetit, Naresh Rajkumar, and Kurt Konolige.
\newblock Going further with point pair features.
\newblock In {\em European conference on computer vision (ECCV)}, pages
  834--848, 2016.

\bibitem{holland1977cauchy}
Paul~W. Holland and Roy~E. Welsch.
\newblock Robust regression using iteratively reweighted least-squares.
\newblock {\em Communications in Statistics - Theory and Methods},
  6(9):813--827, 1977.

\bibitem{huang2017transSync}
Xiangru Huang, Zhenxiao Liang, Chandrajit Bajaj, and Qixing Huang.
\newblock Translation synchronization via truncated least squares.
\newblock In {\em Advances in neural information processing systems (NIPS)},
  pages 1459--1468, 2017.

\bibitem{huang2019learningTransSync}
Xiangru Huang, Zhenxiao Liang, Xiaowei Zhou, Yao Xie, Leonidas~J Guibas, and
  Qixing Huang.
\newblock Learning transformation synchronization.
\newblock In {\em IEEE Conference on Computer Vision and Pattern Recognition
  (CVPR)}, pages 8082--8091, 2019.

\bibitem{huber2003fully}
Daniel~F Huber and Martial Hebert.
\newblock Fully automatic registration of multiple 3d data sets.
\newblock {\em Image and Vision Computing}, 21(7):637--650, 2003.

\bibitem{johnson1999}
A.E. Johnson and M. Hebert.
\newblock Using spin images for efficient object recognition in cluttered 3d
  scenes.
\newblock {\em IEEE Transactions on Pattern Analysis and Machine Intelligence},
  21, 1999.

\bibitem{khoury2017CGF}
Marc Khoury, Qian-Yi Zhou, and Vladlen Koltun.
\newblock Learning compact geometric features.
\newblock In {\em IEEE International Conference on Computer Vision (ICCV)},
  2017.

\bibitem{Kingma2015ADAM}
Diederik~P. Kingma and Jimmy~Lei Ba.
\newblock Adam: a {M}ethod for {S}tochastic {O}ptimization.
\newblock In {\em International Conference on Learning Representations 2015},
  2015.

\bibitem{korman2018latent}
Simon Korman and Roee Litman.
\newblock Latent ransac.
\newblock In {\em IEEE Conference on Computer Vision and Pattern Recognition
  (CVPR)}, pages 6693--6702, 2018.

\bibitem{li20073d}
Hongdong Li and Richard Hartley.
\newblock The 3{D}-3{D} registration problem revisited.
\newblock In {\em International Conference on Computer Vision (ICCV)}, pages
  1--8, 2007.

\bibitem{Lu_2019_DeepVCP}
Weixin Lu, Guowei Wan, Yao Zhou, Xiangyu Fu, Pengfei Yuan, and Shiyu Song.
\newblock Deepvcp: An end-to-end deep neural network for point cloud
  registration.
\newblock In {\em IEEE International Conference on Computer Vision (ICCV)},
  2019.

\bibitem{maset2017practical}
Eleonora Maset, Federica Arrigoni, and Andrea Fusiello.
\newblock Practical and efficient multi-view matching.
\newblock In {\em Proceedings of the IEEE International Conference on Computer
  Vision}, pages 4568--4576, 2017.

\bibitem{mellado2014super}
Nicolas Mellado, Dror Aiger, and Niloy~J Mitra.
\newblock Super 4pcs fast global pointcloud registration via smart indexing.
\newblock In {\em Computer Graphics Forum}, volume~33, 2014.

\bibitem{mian2006three}
Ajmal~S Mian, Mohammed Bennamoun, and Robyn Owens.
\newblock Three-dimensional model-based object recognition and segmentation in
  cluttered scenes.
\newblock {\em IEEE transactions on pattern analysis and machine intelligence},
  28(10):1584--1601, 2006.

\bibitem{moo2018learn2find}
Kwang Moo~Yi, Eduard Trulls, Yuki Ono, Vincent Lepetit, Mathieu Salzmann, and
  Pascal Fua.
\newblock Learning to find good correspondences.
\newblock In {\em IEEE Conference on Computer Vision and Pattern Recognition
  (CVPR)}, pages 2666--2674, 2018.

\bibitem{paszke2017automatic}
Adam Paszke, Sam Gross, Soumith Chintala, Gregory Chanan, Edward Yang, Zachary
  DeVito, Zeming Lin, Alban Desmaison, Luca Antiga, and Adam Lerer.
\newblock Automatic differentiation in {PyTorch}.
\newblock In {\em NIPS Autodiff Workshop}, 2017.

\bibitem{pedregos2011scikit}
F. Pedregosa, G. Varoquaux, A. Gramfort, V. Michel, B. Thirion, O. Grisel, M.
  Blondel, P. Prettenhofer, R. Weiss, V. Dubourg, J. Vanderplas, A. Passos, D.
  Cournapeau, M. Brucher, M. Perrot, and E. Duchesnay.
\newblock Scikit-learn: Machine learning in {P}ython.
\newblock {\em Journal of Machine Learning Research}, 12:2825--2830, 2011.

\bibitem{plotz2018NN3}
Tobias Pl\"{o}tz and Stefan Roth.
\newblock Neural nearest neighbors networks.
\newblock In {\em Advances in Neural Information Processing Systems (NIPS)},
  pages 1087--1098, 2018.

\bibitem{qi2017pointnet}
Charles~R Qi, Hao Su, Kaichun Mo, and Leonidas~J Guibas.
\newblock Pointnet: Deep learning on point sets for 3d classification and
  segmentation.
\newblock In {\em IEEE Conference on Computer Vision and Pattern Recognition
  (CVPR)}, 2017.

\bibitem{rabbani2007}
T. Rabbani, S. Dijkman, F. {van den Heuvel}, and G. Vosselman.
\newblock An integrated approach for modelling and global registration of point
  clouds.
\newblock {\em ISPRS Journal of Photogrammetry and Remote Sensing},
  61:355--370, 2007.

\bibitem{raguram2012usac}
Rahul Raguram, Ondrej Chum, Marc Pollefeys, Jiri Matas, and Jan-Michael Frahm.
\newblock Usac: a universal framework for random sample consensus.
\newblock {\em IEEE transactions on pattern analysis and machine intelligence},
  35(8), 2012.

\bibitem{ranftl2018deepFundamentalMatrix}
Ren{\'e} Ranftl and Vladlen Koltun.
\newblock Deep fundamental matrix estimation.
\newblock In {\em European Conference on Computer Vision (ECCV)}, pages
  284--299, 2018.

\bibitem{ronneberger2015u}
Olaf Ronneberger, Philipp Fischer, and Thomas Brox.
\newblock U-net: Convolutional networks for biomedical image segmentation.
\newblock In {\em International Conference on Medical image computing and
  computer-assisted intervention}, pages 234--241. Springer, 2015.

\bibitem{rusu2009FPFH}
Radu~Bogdan Rusu, Nico Blodow, and Michael Beetz.
\newblock Fast point feature histograms ({FPFH}) for 3{D} registration.
\newblock In {\em IEEE International Conference on Robotics and Automation
  (ICRA)}, 2009.

\bibitem{sorkine2017least}
Olga Sorkine-Hornung and Michael Rabinovich.
\newblock Least-squares rigid motion using svd.
\newblock {\em Computing}, 1(1), 2017.

\bibitem{theiler2015}
Pascal Theiler, Jan~D. Wegner, and Konrad Schindler.
\newblock Globally consistent registration of terrestrial laser scans via graph
  optimization.
\newblock {\em ISPRS Journal of Photogrammetry and Remote Sensing},
  109:126--136, 2015.

\bibitem{theiler2014keypoint}
Pascal~Willy Theiler, Jan~Dirk Wegner, and Konrad Schindler.
\newblock Keypoint-based 4-points congruent sets--automated marker-less
  registration of laser scans.
\newblock {\em ISPRS journal of photogrammetry and remote sensing}, 2014.

\bibitem{tombari2010USC}
Federico Tombari, Samuele Salti, and Luigi Di~Stefano.
\newblock Unique shape context for 3{D} data description.
\newblock In {\em Proceedings of the ACM workshop on 3D object retrieval},
  2010.

\bibitem{tombari2010SHOT}
Federico Tombari, Samuele Salti, and Luigi Di~Stefano.
\newblock Unique signatures of histograms for local surface description.
\newblock In {\em European conference on computer vision (ECCV)}, 2010.

\bibitem{torr1997development}
Philip~HS Torr and David~W Murray.
\newblock The development and comparison of robust methods for estimating the
  fundamental matrix.
\newblock {\em International journal of computer vision}, 24(3):271--300, 1997.

\bibitem{torsello2011multiview}
Andrea Torsello, Emanuele Rodola, and Andrea Albarelli.
\newblock Multiview registration via graph diffusion of dual quaternions.
\newblock In {\em CVPR 2011}, pages 2441--2448. IEEE, 2011.

\bibitem{ulyanov2016instance}
Dmitry Ulyanov, Andrea Vedaldi, and Victor Lempitsky.
\newblock Instance normalization: The missing ingredient for fast stylization.
\newblock {\em arXiv preprint arXiv:1607.08022}, 2016.

\bibitem{Wang_2019_DCP}
Yue Wang and Justin~M. Solomon.
\newblock Deep closest point: Learning representations for point cloud
  registration.
\newblock In {\em The IEEE International Conference on Computer Vision (ICCV)},
  pages 3523--3532, October 2019.

\bibitem{yang2015go}
Jiaolong Yang, Hongdong Li, Dylan Campbell, and Yunde Jia.
\newblock Go-icp: A globally optimal solution to 3d icp point-set registration.
\newblock {\em IEEE transactions on pattern analysis and machine intelligence
  (TPAMI)}, 38(11):2241--2254, 2015.

\bibitem{yew20183dfeatnet}
Zi~Jian Yew and Gim~Hee Lee.
\newblock 3dfeat-net: Weakly supervised local 3d features for point cloud
  registration.
\newblock In {\em European Conference on Computer Vision}, 2018.

\bibitem{yi2016learning}
Kwang~Moo Yi, Yannick Verdie, Pascal Fua, and Vincent Lepetit.
\newblock Learning to assign orientations to feature points.
\newblock In {\em Computer Vision and Pattern Recognition (CVPR)}, 2016.

\bibitem{zeisl2013}
B. Zeisl, K. K\"{o}ser, and M. Pollefeys.
\newblock Automatic registration of rgb-d scans via salient directions.
\newblock In {\em IEEE International Conference on Computer Vision}, pages
  2808--2815, 2013.

\bibitem{zeng20163dmatch}
Andy Zeng, Shuran Song, Matthias Nie{\ss}ner, Matthew Fisher, Jianxiong Xiao,
  and Thomas Funkhouser.
\newblock 3{DM}atch: {L}earning {L}ocal {G}eometric {D}escriptors from {RGB-D}
  {R}econstructions.
\newblock In {\em IEEE Conference on Computer Vision and Pattern Recognition
  (CVPR)}, 2017.

\bibitem{zhang2019oanet}
Jiahui Zhang, Dawei Sun, Zixin Luo, Anbang Yao, Lei Zhou, Tianwei Shen, Yurong
  Chen, Long Quan, and Hongen Liao.
\newblock Learning two-view correspondences and geometry using order-aware
  network.
\newblock In {\em International Conference on Computer Vision (ICCV)}, 2019.

\bibitem{zhou2016FGR}
Qian-Yi Zhou, Jaesik Park, and Vladlen Koltun.
\newblock Fast global registration.
\newblock In {\em European Conference on Computer Vision (ECCV)}, pages
  766--782, 2016.

\bibitem{zhu2018gsfm}
Siyu Zhu, Runze Zhang, Lei Zhou, Tianwei Shen, Tian Fang, Ping Tan, and Long
  Quan.
\newblock Very large-scale global sfm by distributed motion averaging.
\newblock In {\em Proceedings of the IEEE Conference on Computer Vision and
  Pattern Recognition}, pages 4568--4577, 2018.

\end{thebibliography}

}
%}

\setcounter{section}{0}
\renewcommand\thesection{\Alph{section}}
\newcommand{\suppsection}{\subsection}
\section{Supplementary Material}
\section{Supplementary material}
In this supplementary material, we provide additional information about the proposed algorithm~(Sec.~\ref{sec:closed_form_ls_supp}-\ref{sec:closed_form_ts_supp} and Alg.~\ref{alg:3d_mul_reg_supp}), network architectures and training configurations~(Sec.~\ref{sec:network_architecture_supp}), an extended ablation study~(Sec.~\ref{sec:ablation_study_supp}) as well as additional visualizations~(Sec.~\ref{sec:qualitative_supp}). The source code and pretrained models are publicly available under \url{https://github.com/zgojcic/3D_multiview_reg}.
\setcounter{equation}{21}
\vspace{-2ex}
\subsection{Closed-form solution of Eq. 4.}
\label{sec:closed_form_ls_supp}
For the sake of completeness we summarize the closed-form differentiable solution of the weighted least square pairwise registration problem
\begin{equation}
    \hat{\mathbf{R}}_{ij}, \hat{\mathbf{t}}_{ij} = \argmin_{\mathbf{R}_{ij},\mathbf{t}_{ij}} \sum_{l=1}^{N} w_l|| \mathbf{R}_{ij}\mathbf{p}_l + \mathbf{t}_{ij} - \mathbf{q}_l) ||^2.
    \label{eq:weighted_pairwise_supp}
\end{equation}
Let $\overline{\mathbf{p}}$ and $\overline{\mathbf{q}}$ 
\begin{equation}
    \overline{\mathbf{p}} := \frac{\sum^{N_\mathbf{P}}_{l=1} w_l \mathbf{p}_l}{\sum^{N_\mathbf{P}}_{l=1} w_l}, \quad
    \overline{\mathbf{q}} := \frac{\sum^{N_\mathbf{Q}}_{l=1} w_l \mathbf{q}_l}{\sum^{N_\mathbf{Q}}_{l=1} w_l}
\end{equation}
denote weighted centroids of point clouds $\mathbf{P} \in \mathbb{R}^{N \times 3}$ and $\mathbf{Q} \in \mathbb{R}^{N \times 3}$, respectively. The centered point coordinates can then be computed as 
\begin{equation}
    \widetilde{\mathbf{p}}_l := \mathbf{p}_l - \mathbf{\overline{\mathbf{p}}}, \quad \widetilde{\mathbf{q}}_l := \mathbf{q}_l - \mathbf{\overline{\mathbf{q}}}, \quad l = 1,\ldots, N
\end{equation}
Arranging the centered points back to the matrix forms $\widetilde{\mathbf{P}}  \in \mathbb{R}^{N \times 3}$ and $\widetilde{\mathbf{Q}}  \in \mathbb{R}^{N \times 3}$, a weighted covariance matrix $\mathbf{S} \in \mathbb{R}^{3 \times 3}$ can be computed as 
\begin{equation}
    \mathbf{S} = \widetilde{\mathbf{P}}^T\mathbf{W}\widetilde{\mathbf{Q}}
\end{equation}
where $\mathbf{W} = \text{diag}(w_1,\ldots,w_{N})$ . Considering the singular value decomposition $\mathbf{S}=\mathbf{U}\mathbf{\Sigma}\mathbf{V}^T$ the solution to Eq.~\ref{eq:weighted_pairwise_supp} is given by
\begin{equation}
    \hat{\mathbf{R}}_{ij} = \mathbf{V}\begin{bmatrix*}[c] 
    1 & 0 & 0 \\
    0 & 1 & 0 \\
    0 & 0 & \text{det}(\mathbf{V}\mathbf{U}^T)\\ \end{bmatrix*}\mathbf{U}^T  
\end{equation}
where $\text{det}(\cdot)$ denotes computing the determinant and is used here to avoid creating a reflection matrix. Finally, $\hat{\mathbf{t}}_{ij}$ is computed as
\begin{equation}
    \hat{\mathbf{t}}_{ij} = \overline{\mathbf{q}} - \hat{\mathbf{R}}_{ij}\overline{\mathbf{p}}
\end{equation}
\subsection{Closed-form solution of Eq. 5 and 6}
\label{sec:closed_form_ts_supp}
In this section we summarize the closed form solutions to Eq. 5 and 6 from the main paper describing the rotation and translation synchronization, respectively. 

The least squares formulation of the rotation synchronization problem
\begin{equation}
    \mathbf{R}_i^* = \argmin_{\mathbf{R_i} \in SO(3)} \sum_{(i,j) \in \mathcal{E}} c_{ij} ||\hat{\mathbf{R}}_{ij} - \mathbf{R}_i\mathbf{R}_j^T||^2_F
    \label{eq:rot_sync_supp}
\end{equation}
admits a closed form solution under spectral relaxation as follows~\cite{arie2012global, arrigoni2014robust}. Consider a symmetric matrix $\mathbf{L} \in  \mathbb{R}^{3N_\mathcal{S}  \times 3N_\mathcal{S}}$ resembling a block Laplacian matrix, defined as
\begin{equation}
    \mathbf{L} = \mathbf{D} - \mathbf{A}
\end{equation}
where $\mathbf{D}$ is the weighted degree matrix constructed as  
\begin{equation}
\mathbf{D} = \begin{bmatrix*}
    \mathbf{I}_3\sum_i{c_{i1}} & & &\\
    & \mathbf{I}_3\sum_i{c_{i2}} & & \\
    & & \ddots & \\
    & & & \mathbf{I}_3\sum_i{c_{iN_\mathcal{S}}} \\
  \end{bmatrix*}
\end{equation}
and $\mathbf{A}$ is a block matrix of the relative rotations
\begin{equation}
    \mathbf{A}  = \begin{bmatrix*}[c] 
    \mathbf{0}_3 & c_{12}\hat{\mathbf{R}}_{12} & \cdots & c_{1N_\mathcal{S}}\hat{\mathbf{R}}_{1N_\mathcal{S}}\\
    c_{21}\hat{\mathbf{R}}_{21} & \mathbf{0}_3 &  \cdots & c_{2N_\mathcal{S}}\hat{\mathbf{R}}_{2N_\mathcal{S}}\\
    \vdots &  & \ddots & \vdots\\
    c_{N_\mathcal{S}1}\hat{\mathbf{R}}_{N_\mathcal{S}1} &  c_{N_\mathcal{S}2}\hat{\mathbf{R}}_{N_\mathcal{S}2} & \cdots & \mathbf{0}_3 \\
\end{bmatrix*}
\end{equation}
where the weights $c_{ij} := \zeta_{\text{init}}(\bm{\Gamma})$ represent the confidence in the relative transformation parameters $\hat{\mathbf{M}}_{ij}$ and $N_\mathcal{S}$ denotes the number of nodes in the graph. The least squares estimates of the global rotation matrices $\mathbf{R}^*_i$ are then given, under relaxed orthonormality and determinant constraints, by the three eigenvectors $\mathbf{v}_i \in \mathbb{R}^{3N_\mathcal{S}}$ corresponding to the smallest eigenvalues of $\mathbf{L}$. Consequently, the nearest rotation matrices under Frobenius norm can be obtained by a projection of the $3\times3$ submatrices of $\mathbf{V}=[\mathbf{v}_1,\mathbf{v}_2,\mathbf{v}_3] \in \mathbb{R}^{3N_\mathcal{S}\times 3}$ onto the orthonormal matrices and enforcing the determinant $\det(\mathbf{R}_i^*)=1$ to avoid the reflections.

Similarly, the closed-form solution to the least squares formulation of the translation synchronization
\begin{equation}
    \mathbf{t}_i^* = \argmin_\mathbf{t_i} \sum_{(i,j) \in \mathcal{E}} c_{ij}||\hat{\mathbf{R}}_{ij}\mathbf{t}_{i} + \hat{\mathbf{t}}_{ij} - \mathbf{t}_{j} ||^2
    \label{eq:trans_sync_supp}
\end{equation}
can be written as~\cite{huang2019learningTransSync}
\begin{equation}
    \mathbf{t}^* = \mathbf{L}^+\mathbf{b}
\end{equation}
where $\mathbf{t}^* = [\mathbf{t}_1^{*^{T}},\dots,\mathbf{t}_{N_{\mathcal{S}}}^{*^{T}}]^T \in \mathbb{R}^{3N_\mathcal{S}}$ and $\mathbf{b} = [\mathbf{b}_1^{*^{T}},\dots,\mathbf{b}_{N_{\mathcal{S}}}^{*^{T}}]^T \in \mathbb{R}^{3N_\mathcal{S}}$ with

\begin{equation}
    \mathbf{b}_i := -\hspace{-0.1ex}\sum_{j \in \mathcal{N}(i)} c_{ij} \hat{\mathbf{R}}^T_{ij}\hat{\mathbf{t}}_{ij}.
\end{equation}
where $\mathcal{N}(i)$ denotes all the neighboring vertices of $ \mathbf{S}_i$ in graph $\mathcal{G}$.
\begin{figure}[t!]
	\begin{center}
		\includegraphics[width=\columnwidth]{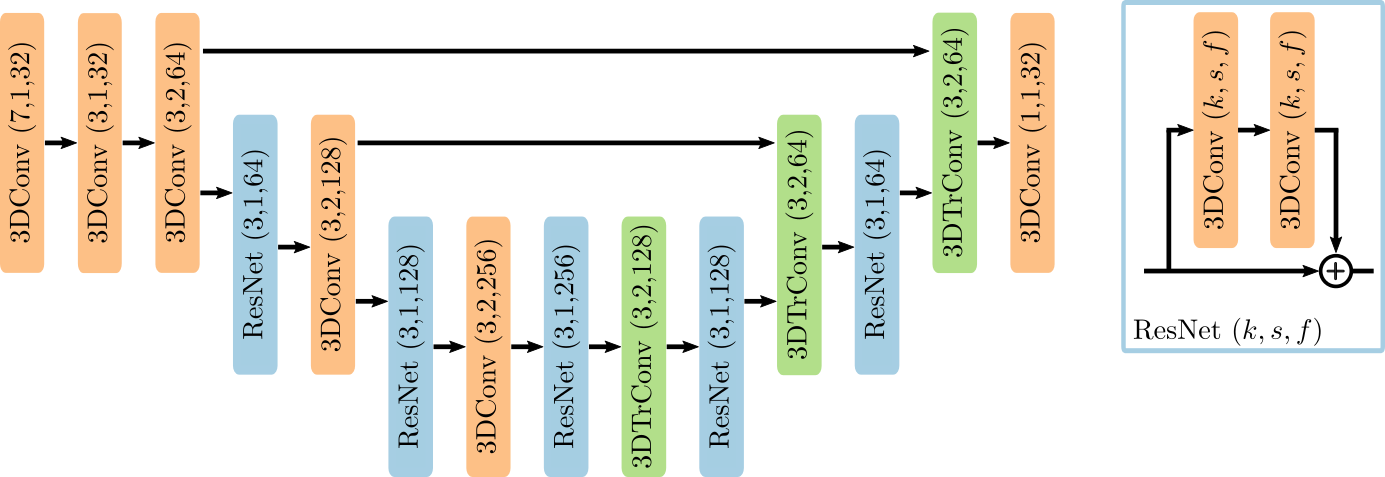}
	\end{center}
	\caption{Network architecture of the FCGF~\cite{Choy2019FCGF} feature descriptor. Each convolutional layer (except the last one) is followed by batch normalization and ReLU activation function.  The numbers in parentheses denote kernel size, stride, and the number of kernels, respectively. }
	\label{fig:FCGF_network}
\end{figure}
\begin{figure*}[!t]
	\begin{center}
		\includegraphics[width=\textwidth]{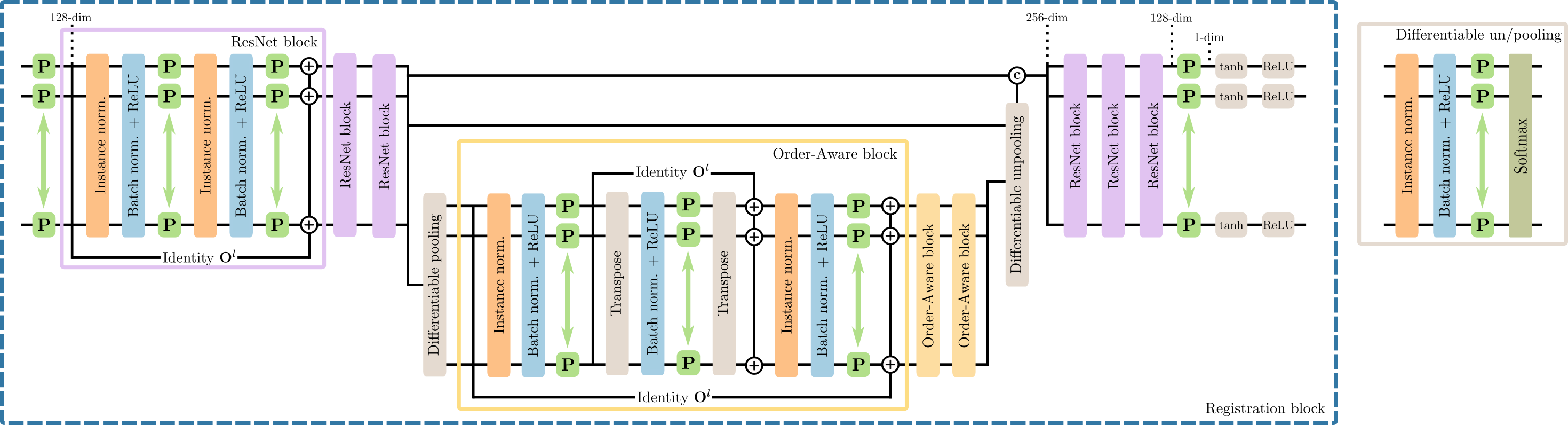}
	\end{center}
	\caption{Network architecture of the registration block consists of two main modules: i) a PointNet-like ResNet block with instance normalization, and ii) an order-aware block. For each point cloud pair, putative correspondences are feed into three consecutive ResNet blocks followed by a differentiable pooling layer, which maps the $N_{c}$ putative correspondences to $M_{c}$ clusters $\mathbf{X}_{k+1}$ at the level $k+1$. These serve as input to the three order-aware blocks. Their output $\mathbf{X}'_{k+1}$ is fed along with $\mathbf{X}_{k}$ into the differentiable unpooling layer. The recovered features are then used as input to the remaining three ResNet blocks. The output of the registration block are the scores $s_i$ indicating whether the putative correspondence is an outlier or an inlier. Additionally, the 128-dim features (denoted as $\mathbf{X}^{\text{conf}} := f_{\theta}^{-2}(\cdot)$) before the last perceptron layer $\mathbf{P}$ are used as input to the confidence estimation block.}
	\label{fig:Reg_block_network}
\end{figure*}
\subsection{Network architecture and training details}
\label{sec:network_architecture_supp}
This section describes the network architecture as well as the training details of the FCGF~\cite{Choy2019FCGF} feature descriptor (Sec.~\ref{sec:FCGF}) and the proposed registration block (Sec.~\ref{sec:reg_block}).
Both networks are implemented in Pytorch and pretrained using the \textit{3DMatch} dataset~\cite{zeng20163dmatch}.

\subsubsection{FCGF local feature descriptor}
\label{sec:FCGF}

\paragraph{Network architecture} The FCGF~\cite{Choy2019FCGF} feature descriptor operates on sparse tensors that represent a point cloud in form of a set of unique coordinates $\mathbf{C}$ and their associated features $\mathbf{F}$

\begin{equation}
    \mathbf{C}  = \begin{bmatrix*}[c] 
    x_1 & y_1 & z_1 & b_1\\
    \vdots &\vdots &\vdots &\vdots\\
    x_N & y_N & z_N & b_N\\ \end{bmatrix*}, \quad\quad
    \mathbf{F}  = \begin{bmatrix*}[c] 
    f_1\\
    \vdots\\
    f_N\\ \end{bmatrix*}\quad
\end{equation}
where $x_i,y_i,z_i$ are the coordinates of the $i$-th point in the point cloud and $f_i$ is the associated feature (in our case simply 1). FCGF is implemented using the Minkowski Engine, an auto-differentiation library, which provides support for sparse convolutions and implements all essential deep learning layers~\cite{choy2019Minkowski}. We adopt the original, fully convolutional network design of FCGF that is depicted in Fig.~\ref{fig:FCGF_network}. It has a UNet structure~\cite{ronneberger2015u} and utilizes skip connections and ResNet blocks~\cite{he2016deep} to extract the per-point $32$ dim feature descriptors. To obtain the unique coordinates $\mathbf{C}$, we use a GPU implementation of the voxel grid downsampling~\cite{choy2019Minkowski} with the voxel size $v := 2.5\,\text{cm}$.
\paragraph{Training details}
We again follow~\cite{Choy2019FCGF} and pre-train FCGF for 100 epochs using the point cloud fragments from the \textit{3DMatch} dataset~\cite{zeng20163dmatch}. We optimize the parameters of the network using stohastic gradient descent with a batch size $4$ and an initial learning rate of $0.1$ combined with an exponential decay with $\gamma = 0.99$. To introduce rotation invariance of the descriptors we perform a data augmentation by randomly rotating each of the fragments along an arbitrary direction, by a different rotation, sampled from the $[0^\circ,360^\circ)$ interval. The sampling of the positive and negative examples follows the procedure proposed in~\cite{Choy2019FCGF}.
\subsubsection{Registration block}
\label{sec:reg_block_supp}
\paragraph{Network architecture}
The architecture of the registration block (same for $\psi_{\text{init}}(\cdot)$ and $\psi_{\text{iter}}(\cdot)$)\footnote{For $\psi_{\text{init}}(\cdot)$ the input dimension is increased from 6 to 8 (weights and residuals added).} follows~\cite{zhang2019oanet} and is based on the PointNet-like architecture~\cite{qi2017pointnet} where each of the fully connected layers ($\mathbf{P}$ in Fig.~\ref{fig:Reg_block_network}) operates on individual correspondences. The local context is then aggregated using the instance normalization layers~\cite{ulyanov2016instance} defined as 
\begin{equation}
    \mathbf{y}_i^l = \frac{\mathbf{x}_i^l- \boldsymbol{{\mu}}^l}{\boldsymbol{\sigma}_l}
\end{equation}
where $\mathbf{x}_i^l$ is the output of the layer $l$ and $\boldsymbol{{\mu}}^l$ and $\boldsymbol{\sigma}_l$ are per dimension mean value and standard deviation, respectively. Opposed to the more commonly used batch normalization, instance normalization operates on individual training examples and not on the whole batch. Additionally, to reinforce the local context, the order-aware blocks~\cite{zhang2019oanet} are used to map the correspondences to clusters using the learned soft pooling $\mathbf{S}_{\text{pool}} \in \mathbb{R}^{N_c \times M_c}$ and unpooling $\mathbf{S}_{\text{unpool}} \in \mathbb{R}^{N_c \times M_c}$ operators as
\begin{equation}
    \mathbf{X}_{k+1} = \mathbf{S}^T_{\text{pool}} \mathbf{X}_k \quad \text{and} \quad \mathbf{X}'_k = \mathbf{S}_{\text{unpool}}\mathbf{X}_{k+1}'
\end{equation}
where $N_c$ is the number of correspondences and $M_c$ is the number of clusters. $\mathbf{X}_k$ and $\mathbf{X}_{k+1}$ are the features at the level $k$ (before clustering) and $k+1$ (after clustering), respectively (see Fig.~\ref{fig:Reg_block_network}). Finally, $\mathbf{X}'_{k+1}$ denotes the output of the last layer in the level $k+1$.

\paragraph{Training details}
We pre-train the registration blocks using the same fragments from the \textit{3DMatch}~dataset. Specifically, we first infer the FCGF descriptors and randomly sample $N_c=5000$ descriptors per fragment. We use these descriptors to compute the putative correspondences for all fragment pairs $(i,j)$ such that $i \leq j$. Based on the ground truth transformation parameters, we label these correspondences as inliers if the Euclidean distance between the points after the transformation is smaller than $7.5$~cm. At the start of the training (first 15000 iterations) we supervise the learning using only the binary cross-entropy loss. Once a meaningful number of correspondences can already be classified correctly we add the transformation loss. We train the network for $500$k iterations using Adam~\cite{Kingma2015ADAM} optimizer with the initial learning rate of $0.001$. We decay the learning rate every $1000$ iterations by multiplying it with $0.999$. To learn the rotation invariance we perform data augmentation, starting from the $25000$th iteration, by randomly sampling an angle from the interval $[0^\circ,n_a\cdot20^\circ)$ where $n_a$ is initialized with zero and is then increased by $1$ every $5000$ iteration until the interval becomes $[0^\circ,360^\circ)$.
\subsection{Pseudo-code}
Alg.~\ref{alg:3d_mul_reg_supp} shows the pseudo-code of our proposed approach. We iterate $k=4$ times over the network and transformation synchronization (i.e.~\textit{Transf-Sync}) layers and in each of those iterations we execute the \textit{Transf-Sync} layer four times. Our implementation is constructed in a modular way (each part can be run on its own) and can accept a varying number of input point clouds with or without the connectivity information. 
\begin{algorithm}[!t]
	\caption{Pseudo-code of the proposed approach}\label{alg:3d_mul_reg_supp}
	\small{\begin{algorithmic}
			\State{Input: a set of potentially overlapping scans $\{\mathbf{S}_i\}_{i=1}^{N_{\mathcal{S}}}$}
			\State{Output: globally optimized poses $\{\mathbf{M}_i^{*}\}_{i=1}^{N_{\mathcal{S}}}$}
			\State{\textcolor{bostonuniversityred}{\textit{\# Compute the pairwise transformations}}}\\
			\For{each pair of scans $\mathbf{S}_i, \mathbf{S}_j \subset \mathcal{S}, i \neq j$}{
				\State{\textcolor{bostonuniversityred}{\# find the putative correspondences using $\phi(\cdot)$}}
				\State{\text{- }$\mathbf{X}_{ij}=\operatorname{cat}([\mathbf{S}_i, \phi(\mathbf{S}_i,\mathbf{S}_j)])\in\mathbb{R}^{N_{\mathbf{S}_i}\times 6}$}
				\State{\textcolor{bostonuniversityred}{\# compute the weights $\mathbf{w}_{ij}\in\mathbb{R}^{n_{\mathbf{S}_i}}$ using $\psi_{init}(\cdot)$}}
				\State {\text{- }$\mathbf{w}_{ij}=\psi_{init}(\mathbf{X}_{ij})\in\mathbb{R}^{N_{\mathbf{S}_i}}$}
				\State{\text{- }calculate $\mathbf{R}_{ij}, \mathbf{t}_{ij}$ using SVD according to (4)} %
			}
			\State{\textcolor{bostonuniversityred}{\textit{\# Iterative network for transformation synchronization}}}
			\State{$\mathbf{X}_{ij}^{(0)}\leftarrow \mathbf{X}_{ij}$, $\mathbf{w}_{ij}^{(0)}\leftarrow \mathbf{w}_{ij}$, $\mathbf{r}_{ij}^{(0)}\leftarrow \mathbf{r}_{ij}$}\\
			\For{$k=1,2,\cdots,\mathrm{max\_iters}$}{
				\hspace{-2ex}\For{each pairwise output from $\psi_{init}$}{
					\State{\text{- }$\mathbf{R}_{ij}^{(k)}, \mathbf{t}_{ij}^{(k)}, \mathbf{w}_{ij}^{(k)}=\psi_{iter}([\mathbf{X}_{ij}^{(k-1)}, \mathbf{w}_{ij}^{(k-1)}, \mathbf{r}_{ij}^{(k-1)}])$}
					\State{\text{- }estimate $\textit{local}\{c^{(k)}_{ij}\}$ using (16)}%
				}
				\State{\text{- }Gather the pairwise estimation as $\mathbf{R}^{(k)}, \mathbf{t}^{(k)}, \mathbf{c}^{(k)}$} %
				\State{\textcolor{bostonuniversityred}{\# Build the graph and perform the synchronization}}\\
				\If{$k=1$}{
					\hspace{-4ex}\text{- }$\mathbf{c}^{(k)}:= \textit{local}\{\mathbf{c}^{(k)}\}$
				}
				\hspace{-4ex}\Else
				{
					\hspace{-4ex}\text{- }$\mathbf{c}^{(k)}:= {f}_{HM}(\textit{local}\{\mathbf{c}^{(k)}\} , \textit{global}\{\mathbf{c}^{(k-1)}\}$
				}
				\State{\text{- }$\mathbf{R}^{*(k)}, \mathbf{t}^{*(k)} = \operatorname{Transf\text{-}Sync}(\mathbf{R}^{(k)}, \mathbf{t}^{(k)},\mathbf{c}^{(k)})$}
				\State{\textcolor{bostonuniversityred}{\# update step}}\\
				\For{each pair of scans $\mathbf{S}_i, \mathbf{S}_j \subset \mathcal{S}, i \neq j$}{
					\State{\text{- }$\mathbf{X}_{ij}^{(k+1)}=\operatorname{cat}([\mathbf{S}_i, \mathbf{M}_{ij}^{*(k)}\otimes\phi(\mathbf{S}_i,\mathbf{S}_j)]$}
					\State{\text{- }$\mathbf{w}_{ij}^{(k+1)}=\mathbf{w}_{ij}^{(k)}$}
					\State{\text{- }$\mathbf{r}_{ij}^{(k+1)}=\|\mathbf{S}_i - \mathbf{M}_{ij}^{*(k)}\otimes\phi(\mathbf{S}_i,\mathbf{S}_j)\|_{2}$}
				}
			}
	\end{algorithmic}}
\end{algorithm}
\subsection{Extended ablation study}
\label{sec:ablation_study_supp}
We extend the ablation study presented in the main paper, by analyzing the impact of edge pruning based on the local confidence (i.e.~the output of the confidence estimation block) (Sec.~\ref{sec:edge_pruning}) and of the weighting scheme~(Sec.~\ref{sec:weighting_scheme}) on the angular and translation errors. The ablation study is performed on the point cloud fragments of the \textit{ScanNet} dataset~\cite{dai2017scannet}. %
\subsubsection{Impact of the edge pruning threshold}
\label{sec:edge_pruning}
\begin{figure*}[t!]
    \centering
    \subfloat[]{\includegraphics[width=\columnwidth]{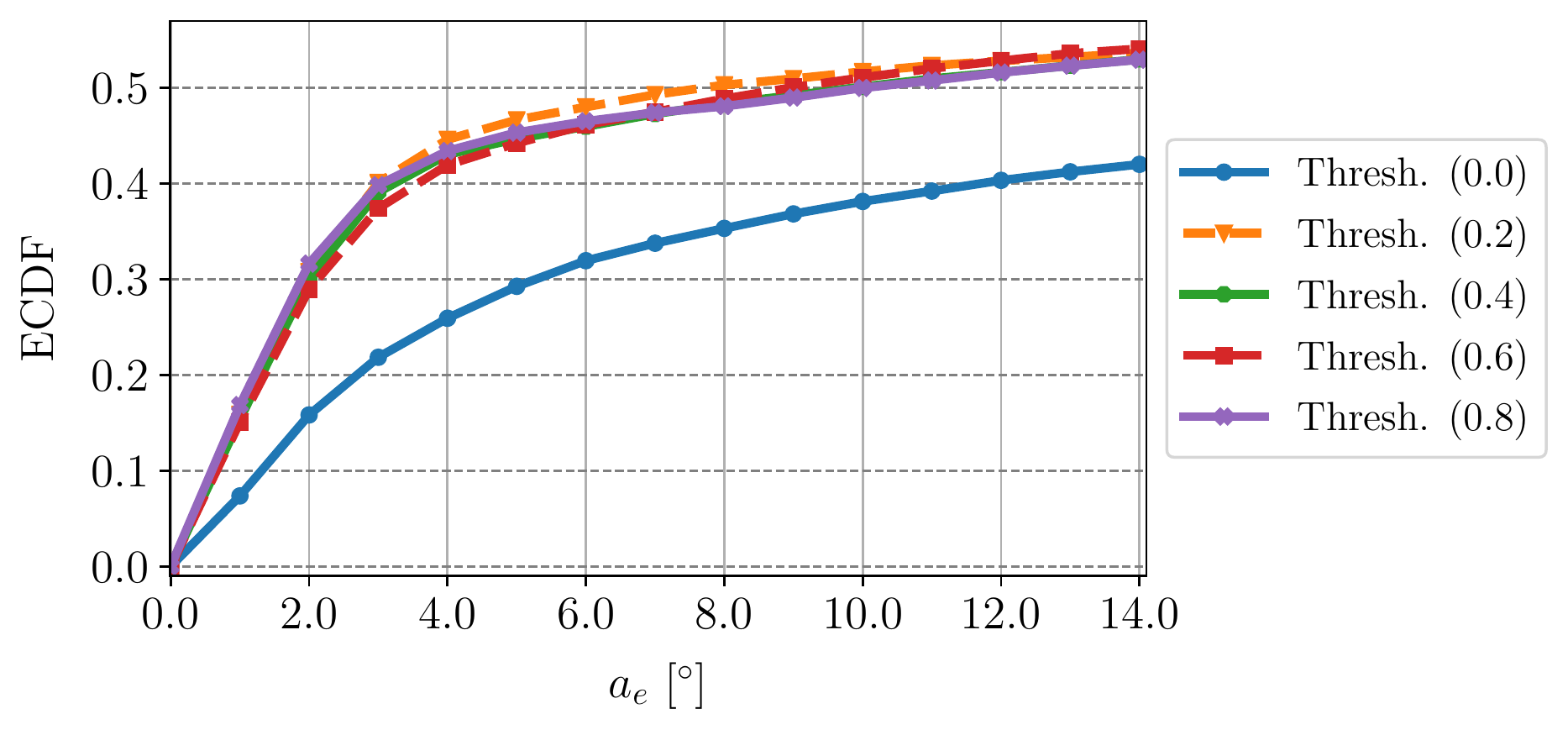}
    \label{subfig:thresh_ae}}\hfill
    \subfloat[]{\includegraphics[width=\columnwidth]{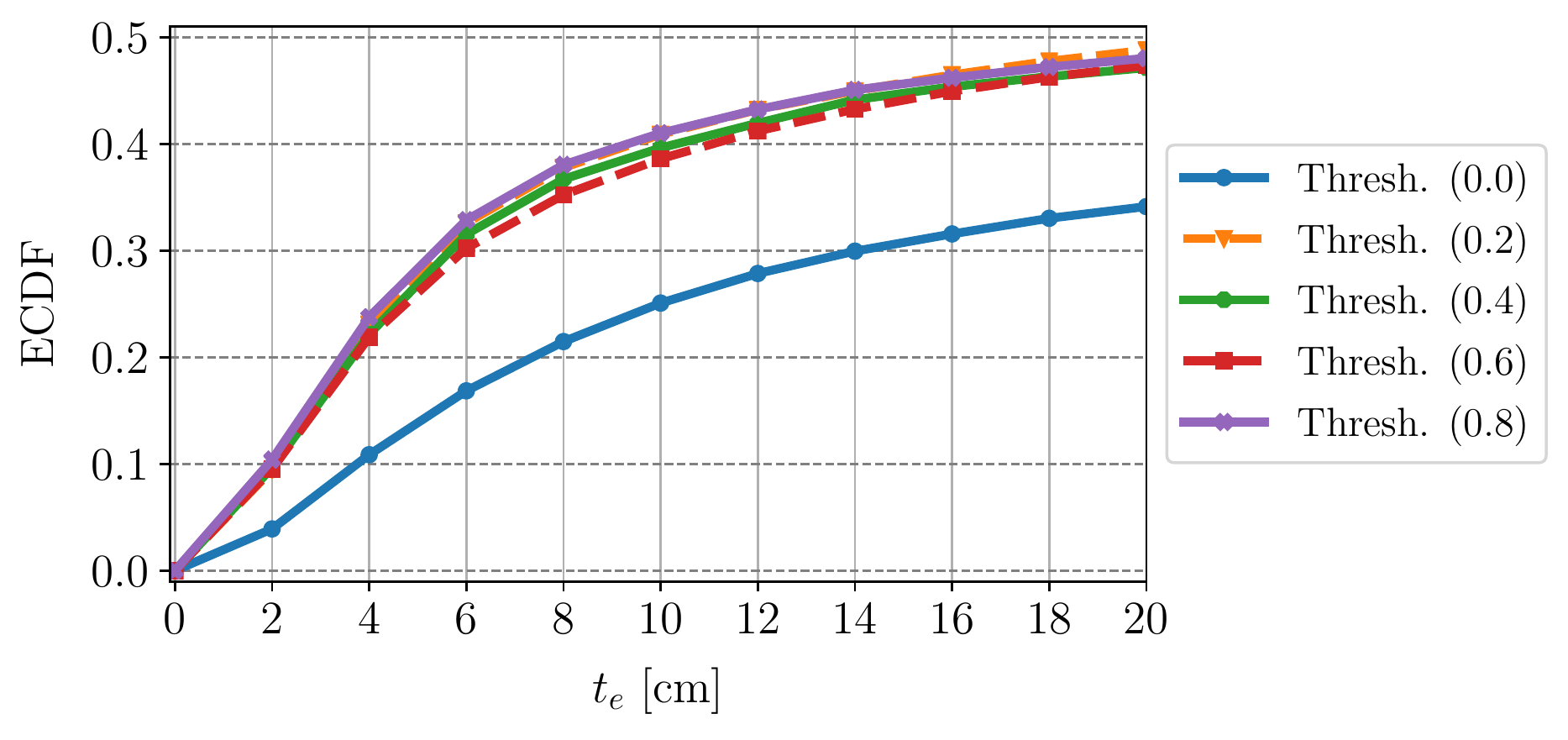}
    \label{subfig:thresh_te}}
    \caption{Impact of the threshold value for edge pruning on the angular and translation errors. Results are obtained using the all pairs as input graph on \textit{ScanNet} dataset~\cite{dai2017scannet}. (a) angular error and (b) translation error.}
    \label{fig:thresh_ae_te}
\end{figure*}

Results depicted in Fig.~\ref{fig:thresh_ae_te} show that the threshold value used for edge pruning has little impact on the angular and translation errors as long as it is larger than 0.2.
\subsection{Impact of the harmonic mean weighting scheme}
\label{sec:weighting_scheme}
In this work, we have introduced a scheme for combining the local and global confidence using the harmonic mean (HM). In the following, we perform the analysis of this proposal and compare its performance to established methods based only on global information~\cite{arrigoni2016se3sync}. To this end, we again consider the scenario "Ours~(Good)" as the input graph connectivity information. We compare the results of the proposed scheme (HM) to SE3 EIG~\cite{arrigoni2016se3sync}, which proposes using the Cauchy function for computing the global edge confidence~\cite{arrigoni2016se3sync}. Note, we use the same pairwise transformation parameters, estimated using the method proposed herein, for all methods.
\begin{figure*}[t!]
    \centering
    \subfloat[]{\includegraphics[width=\columnwidth]{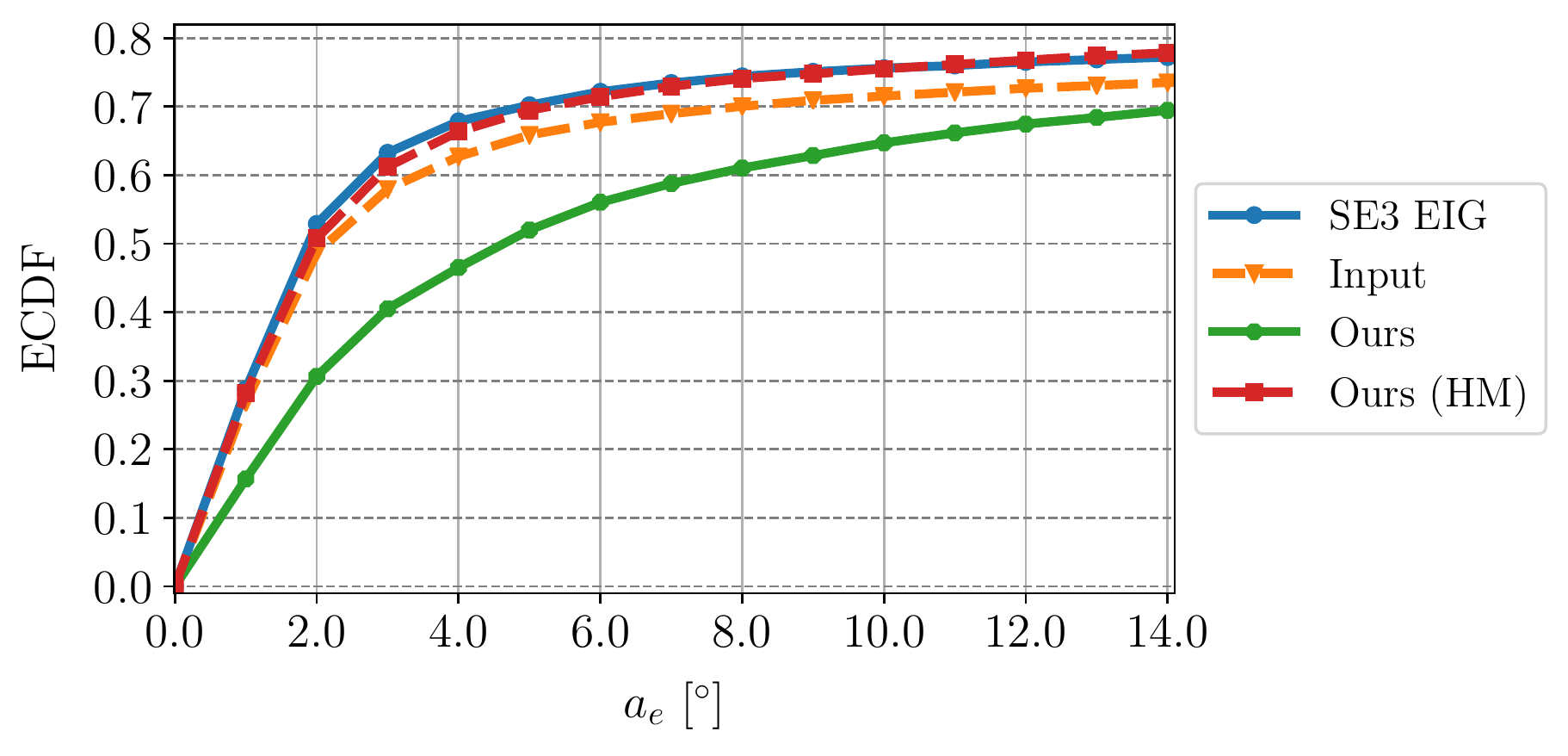}
    \label{subfig:weighting_woec_ae}}\hfill
    \subfloat[]{\includegraphics[width=\columnwidth]{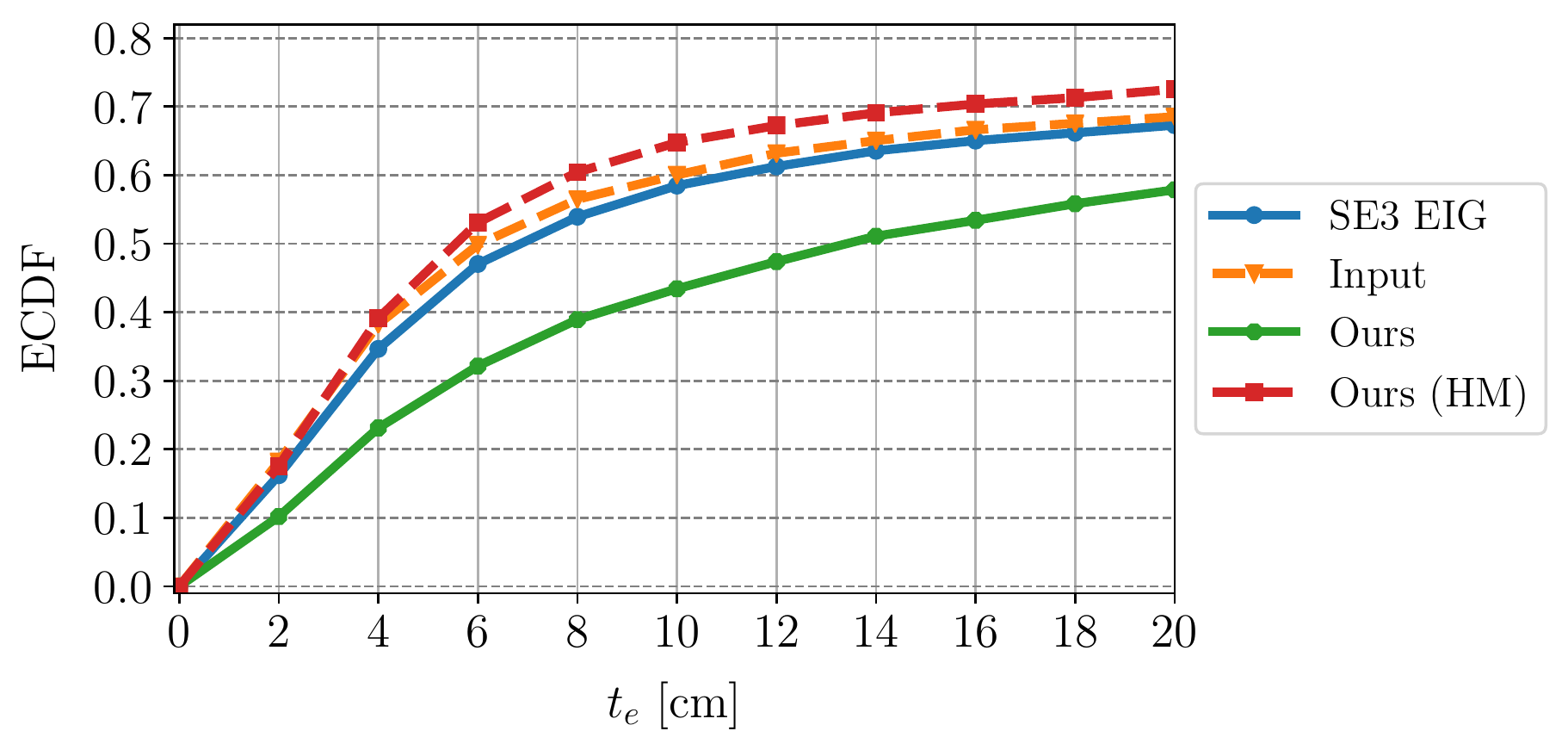}
    \label{subfig:weighting_woec_te}}
    \caption{Impact of the weighting scheme without edge cutting on the angular and translation errors. (a) angular and (b) translation errors.}
    \label{fig:weighting_woec}
\end{figure*}
\begin{figure*}[t!]
    \centering
    \subfloat[]{\includegraphics[width=\columnwidth]{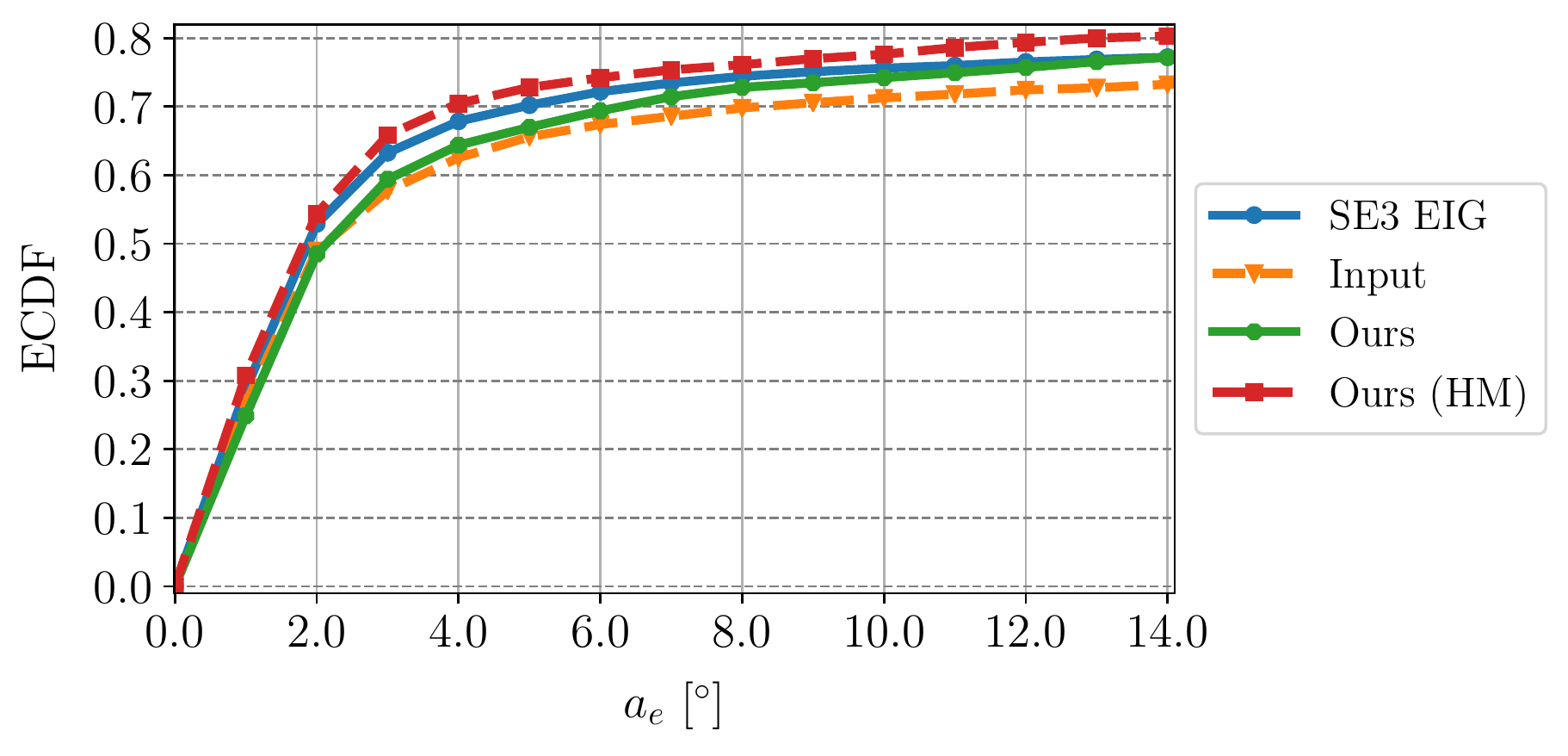}
    \label{subfig:weighting_wec_ae}}\hfill
    \subfloat[]{\includegraphics[width=\columnwidth]{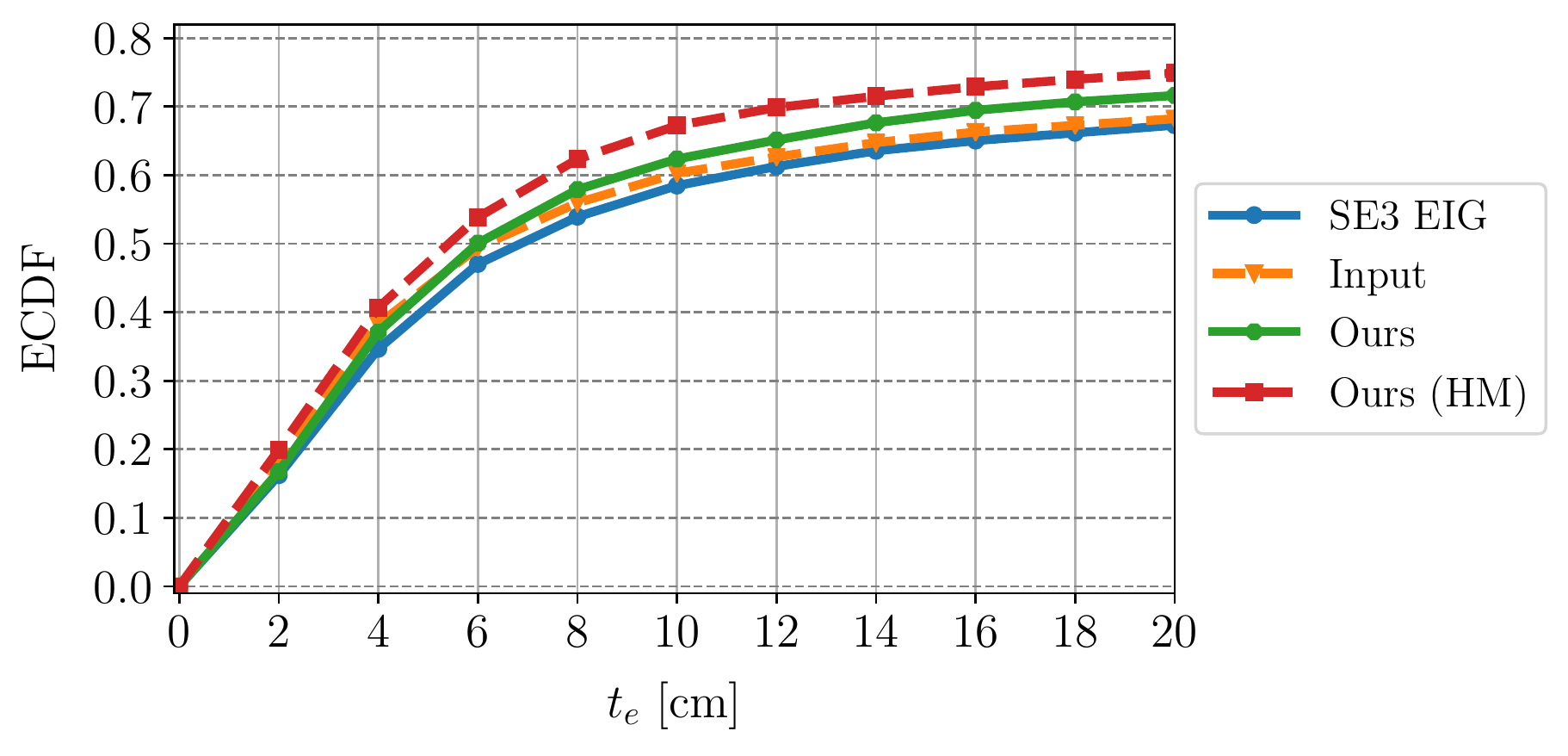}
    \label{subfig:weighting_wec_te}}
    \caption{Impact of the weighting scheme combined with edge cutting, on the angular and translation errors. (a) angular error and (b) translation error.}
    \label{fig:weighting_wec}
\end{figure*}

\paragraph{Without edge pruning}
It turns out that combining the local and global evidence about the graph connectivity is essential to achieve good performance.  In fact, merely relying on local confidence estimates without HM weighting (denoted as ours; green) in Fig.~\ref{fig:weighting_woec}) the \textit{Transf-Sync} is unable to recover global transformations from the given graph connectivity evidence that is very noisy. %
Introducing the HM weighting scheme allows us to reduce the impact of noisy graph connectivity built solely using local confidence and can significantly improve performance after \textit{Transf-Sync} block, which in turn enables us to outperform the \textit{SE3 EIG}.

\paragraph{With edge pruning}
Fig.~\ref{fig:weighting_wec} shows that pruning the edges can help coping with noisy input graph connectivity built from the pairwise input. In principal, suppression of the edges with low confidence results in discarding the outliers that corrupt the l2 solution and as a result improves the performance of the \textit{Transf-Sync} block. %

\subsection{Qualitative results}
\label{sec:qualitative_supp}
\begin{figure*}[!t]
	\begin{center}
		\includegraphics[width=0.9\textwidth]{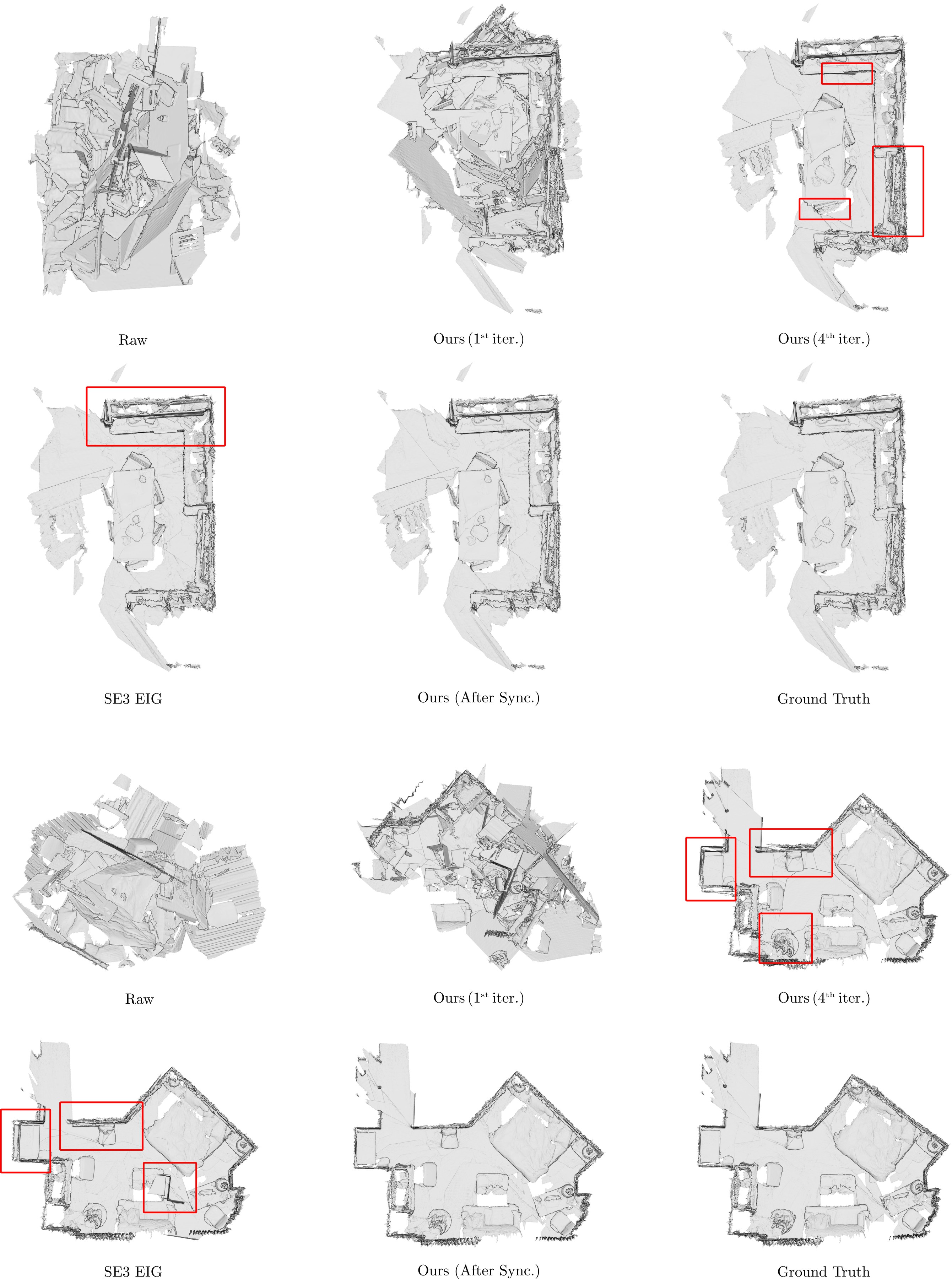}
	\end{center}
	\caption{Selected \textbf{success cases} of our method on \textit{3DMatch} dataset. Top: \textbf{Kitchen} and bottom: \textbf{Hotel 1}. Red rectangles highlight interesting areas with subtle changes.}
	\label{fig:3DMatch_good}
\end{figure*}

\begin{figure*}[!t]
	\begin{center}
		\includegraphics[width=0.92\textwidth]{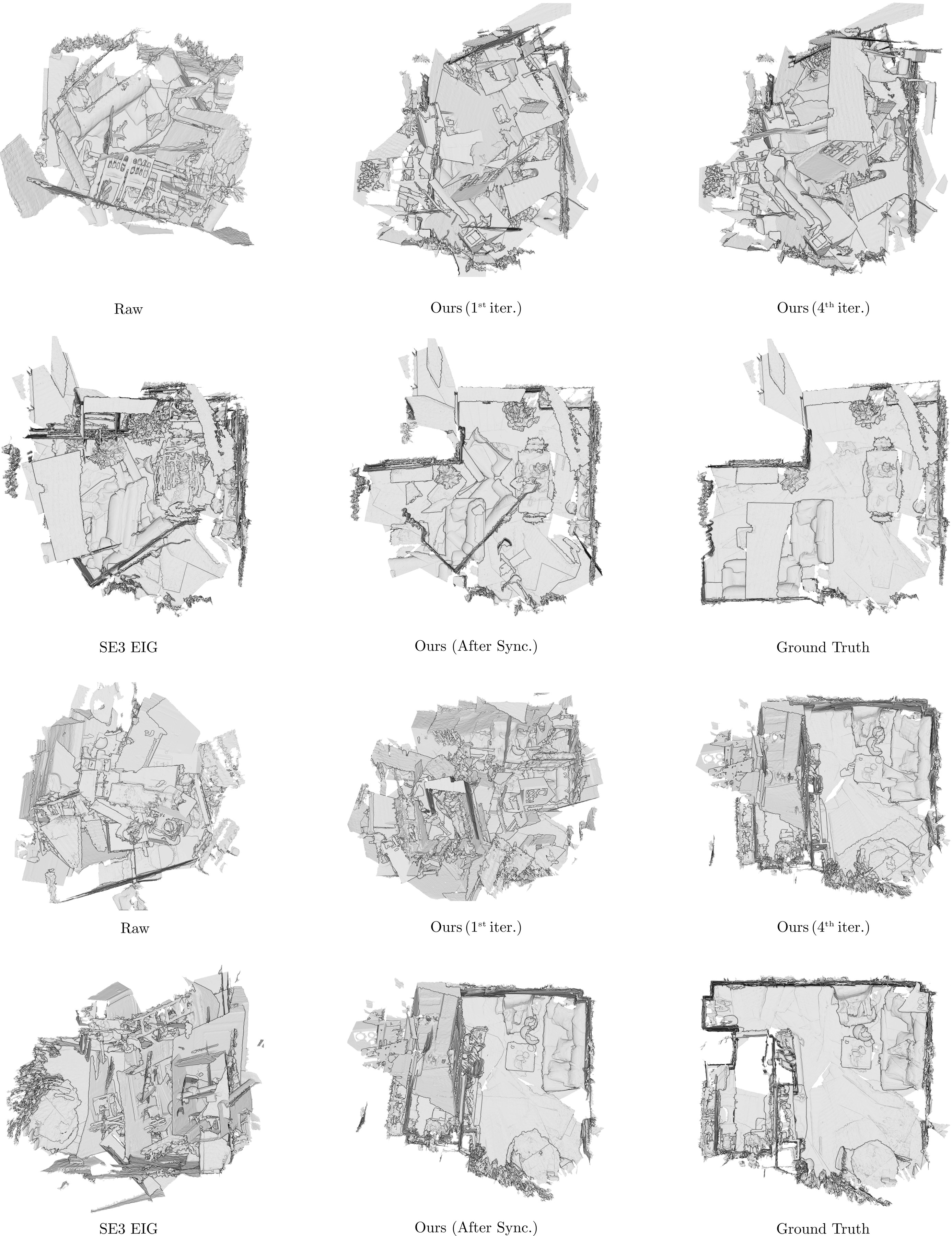}
	\end{center}
	\caption{Selected \textbf{failure cases} of our method on \textit{3DMatch} dataset. Top: \textbf{Home 1} and bottom: \textbf{Home 2}. Note that our method still provides qualitatively better results than state-of-the-art.}
	\label{fig:3DMatch_bad}
\end{figure*}

\begin{figure*}[!t]
	\begin{center}
		\includegraphics[width=0.9\textwidth]{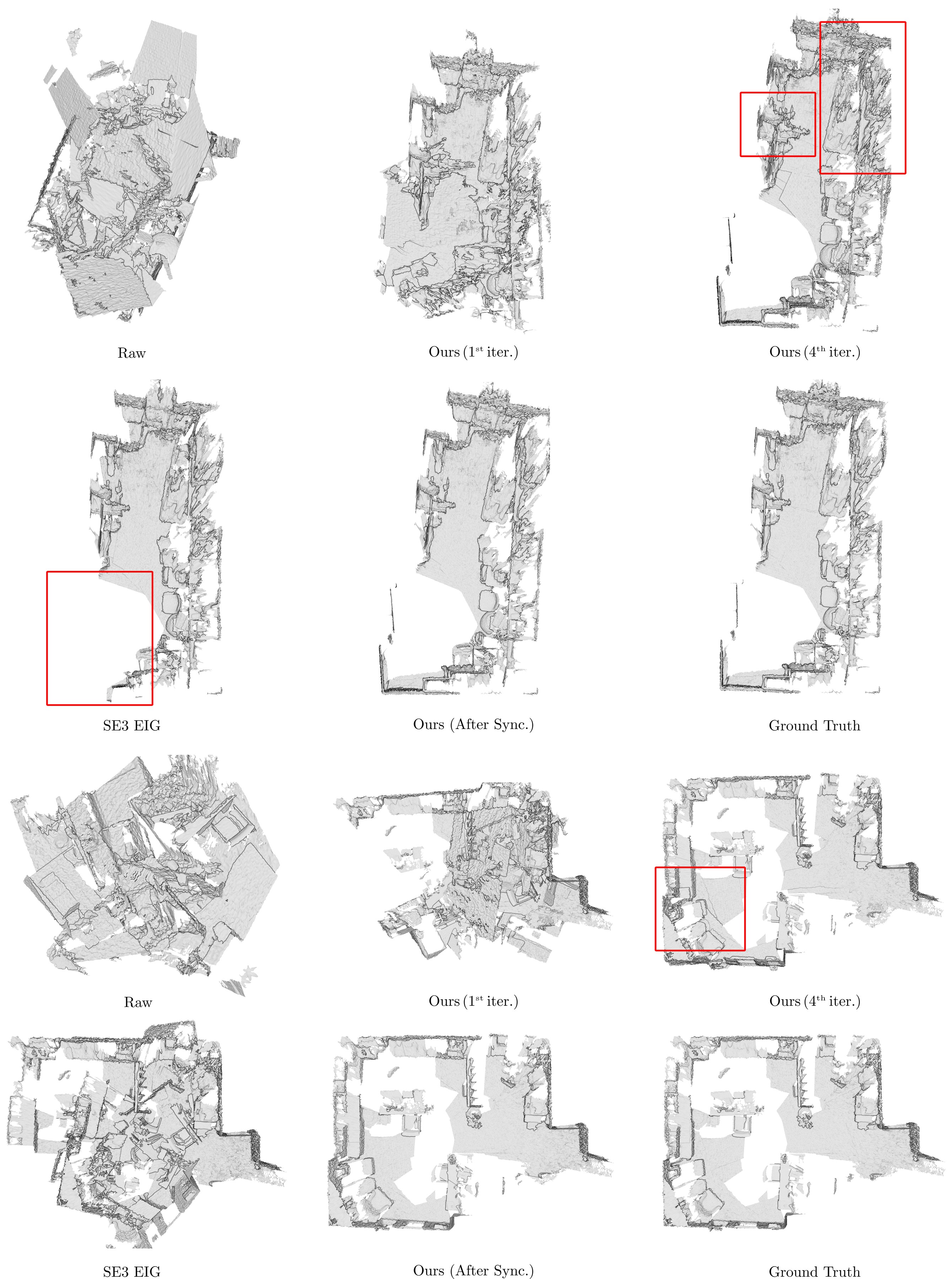}
	\end{center}
	\caption{Selected \textbf{success cases} of our method on \textit{ScanNet} dataset. Top: \textbf{scene0057\_01} and bottom: \textbf{scene0309\_00}. Red rectangles highlight interesting areas with subtle changes.}
	\label{fig:ScanNet_good}
\end{figure*}

\begin{figure*}[!t]
	\begin{center}
		\includegraphics[width=0.93\textwidth]{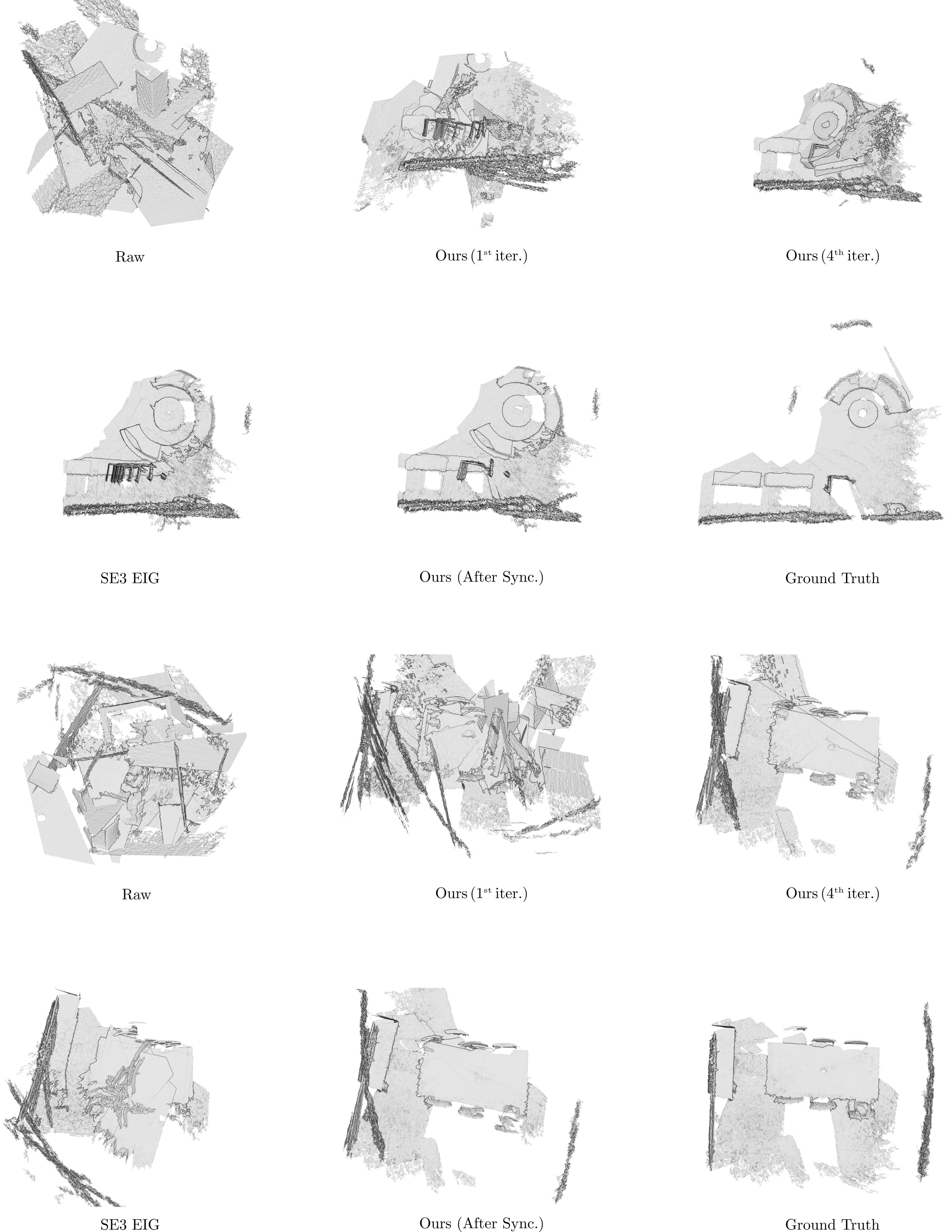}
	\end{center}
	\caption{Selected \textbf{failure cases} of our method on \textit{ScanNet} dataset. Top: \textbf{scene0334\_02} and bottom: \textbf{scene0493\_01}. Note that our method still provides qualitatively better results than state-of-the-art.}
	\label{fig:ScanNet_bad}
\end{figure*}

We provide some additional qualitative results in form of success and failure cases on selected scenes of \textit{3DMatch} (Fig. \ref{fig:3DMatch_good} and \ref{fig:3DMatch_bad}) and \textit{ScanNet}~(Fig.~\ref{fig:ScanNet_good} and \ref{fig:ScanNet_bad}) datasets. Specifically, we compare the results of our whole pipeline \textit{Ours (After Sync.)} to the results of \textit{SE3 EIG}~\cite{arrigoni2016se3sync}, pairwise registration results of our method from the first iteration \textit{Ours ($1^{st}$ iter.)}, and pairwise registration results of our method from the fourth iteration \textit{Ours ($4^{th}$ iter.)}. Both global methods (\textit{Ours (After Sync.)} and \textit{SE3 EIG}) use transformation parameters estimated by our proposed pairwise registration algorithm as input to the transformation synchronization. The failure cases of our method predominantly occur on point clouds with low level of structure (planar areas in Fig.~\ref{fig:3DMatch_bad} bottom) or high level of symmetry and repetitive structures (Fig.~\ref{fig:ScanNet_bad} top and bottom, respectively).


\end{document}